%% file: main.tex
\documentclass{article}

\PassOptionsToPackage{numbers}{natbib}
\usepackage[preprint]{neurips_2026}
\usepackage{subcaption}
\usepackage{hyperref}
\usepackage{wrapfig,adjustbox}

\usepackage[table, dvipsnames, svgnames]{xcolor}
\usepackage{algorithm,algorithmic}

\usepackage[utf8]{inputenc} 
\usepackage[T1]{fontenc}    
\usepackage{url}            
\usepackage{booktabs}       
\usepackage{amsfonts}       
\usepackage{nicefrac}       
\usepackage{microtype}      
\usepackage{xcolor}         

\usepackage{graphicx}
\usepackage{amsmath}
\usepackage{amssymb}
\usepackage{mathtools}
\usepackage{amsthm}
\usepackage[capitalize,noabbrev]{cleveref}

\usepackage{euscript,mdframed}
\usepackage{microtype,relsize}
\usepackage{makecell}
\usepackage{multirow}
\usepackage{booktabs}
\usepackage{bm}
\usepackage{threeparttable}

\usepackage{tcolorbox}

\usepackage{bbm}
\usepackage{xspace}

\theoremstyle{plain}

\theoremstyle{definition}

\theoremstyle{remark}

\definecolor{myblue}{HTML}{096C6C}

\newtcolorbox[auto counter,number within=section]{pabox}[2][]{colback=gray!5!white,colframe=gray!75!black,fonttitle=\bfseries,boxrule=2pt,
title=Prompt~\thetcbcounter: #2,#1}
\newtcolorbox{messagebox}[2][]{colback=gray!5!white,colframe=gray!75!black,fonttitle=\bfseries,boxrule=1pt,
title=#2,
}
\usepackage{fancybox}
\usepackage{enumitem}
\usepackage{amssymb}

\title{Joint Consistency: A Unified Test-Time Aggregation Framework via Energy Minimization}
\hypersetup{
  hidelinks,
  pdftitle={Joint Consistency: A Unified Test-Time Aggregation Framework via Energy Minimization},
  pdfauthor={Yunzhen Yao, Hongye Wang, Yahong Wang, Michael C. Gastpar, Bo Jiang, Lie He}
}

\author{%
  \normalfont
  \begin{tabular}{@{}ccc@{}}
  \makebox[0.30\textwidth][c]{\textbf{Yunzhen Yao}\thanks{Equal contribution.}} &
  \makebox[0.30\textwidth][c]{\textbf{Hongye Wang}\footnotemark[1]} &
  \makebox[0.30\textwidth][c]{\textbf{Yahong Wang}} \\
  EPFL & SUFE & SUFE \\
  {\footnotesize\texttt{yunzhen.yao@epfl.ch}} &
  {\footnotesize\texttt{ishongyewang@gmail.com}} &
  {\footnotesize\texttt{2025214412@stu.sufe.edu.cn}} \\[1.5ex]
  \textbf{Michael C. Gastpar} &
  \textbf{Bo Jiang}\thanks{Corresponding authors.} &
  \textbf{Lie He}\footnotemark[2] \\
  EPFL & SUFE & SUFE\thanks{MoE Key Laboratory of Interdisciplinary Research of Computation and Economics, Shanghai University of Finance and Economics (SUFE).} \\
  {\footnotesize\texttt{michael.gastpar@epfl.ch}} &
  {\footnotesize\texttt{syebojiang@gmail.com}} &
  {\footnotesize\texttt{helie@sufe.edu.cn}}
  \end{tabular}
}

\begin{document}

\maketitle

\begin{abstract}
This paper studies test-time aggregation, an approach that generates multiple reasoning traces and aggregates them into a final answer. Most existing methods rely on evaluation signals collected from candidate traces in isolation or answer frequencies, while ignoring comparative interactions among candidates. We propose \emph{Joint Consistency} (JC), formulated as a constrained Ising-type energy minimization problem, which employs both independent evaluation signals and pairwise comparisons using LLM-as-a-judge. JC provides a unified framework for test-time aggregation that subsumes typical voting and weighted aggregation methods as special cases. Moreover, we develop an efficient approximation strategy that makes interaction modeling practical for large-scale test-time aggregation. Experiments on math and code reasoning benchmarks show that JC outperforms existing baselines across tasks, judge models, trace budgets, and trace-generation settings, with particularly pronounced gains on challenging problems.
\end{abstract}

\section{Introduction}
Scaling test-time computation is an effective approach for improving the performance of large language models (LLMs) on complex reasoning tasks. 
Given a specific query, Test-Time Scaling (TTS) methods allocate additional computation at inference time to improve performance. This is typically achieved by eliciting additional reasoning steps within a single response~\citep{wei2022chain} or by aggregating multiple candidate responses into a final decision~\citep{snell2024scaling,welleck2024decoding,balachandran2025inference}. This additional computation can be executed in parallel~\citep{wang2023self} or sequentially~\citep{madaan2023self, qu2024recursive}, and may incorporate evaluation signals at different stages of the reasoning process~\citep{yao2023tree,fu2025deep}.

For a wide range of TTS methods, the central question is 
\textit{how to effectively aggregate diverse candidate reasoning traces into a single, reliable output}. We formulate this post-hoc decision problem as \textit{Test-Time Aggregation} (TTA).
Existing approaches to TTA include Self-Consistency~\citep{wang2023self}, Weighted Self-Consistency~\citep{guo2025reward}, Best-of-$N$ selection~\citep{jinnai2025regularized}, Self-Certainty~\citep{kang2025scalable}, and a broad class of evaluation-based TTS methods~\citep{cobbe2021training}, where candidate solutions are first fully generated and subsequently evaluated, compared, and/or aggregated.

Most TTA methods evaluate reasoning traces \emph{in isolation}:
the score of each candidate is \emph{independent} of others, 
disregarding comparative information across traces. 
This requires the evaluator to produce well-calibrated absolute quality scores, which are often difficult to obtain.
In contrast, comparative judgments between candidate traces can be more reliable, as they focus on \emph{relative} quality rather than \emph{absolute} scores, a principle well-established in preference learning and LLM alignment~\citep{christiano2017deep,ouyang2022training}.

Nevertheless, a key question remains open: how can we incorporate pairwise relations (i.e., comparative judgments) into test-time aggregation in a principled and computationally efficient manner?
It requires us to design a principled aggregation rule that leverages multiple comparative judgments, and make it scalable beyond the naive $O(N^2)$ cost of collecting all-pair comparisons.

\paragraph{Contributions.} 
To address the above challenges, we propose \textit{Joint Consistency} (JC) for Test-Time Aggregation (TTA), formulated as a \emph{constrained energy minimization problem} with an Ising-type structure. 
Our contributions are threefold: 
\begin{itemize}
    
    \item \textbf{Modeling pairwise interactions in test-time aggregation.}
    Our proposed constrained energy minimization model enables JC to incorporate pairwise comparative signals through an \emph{interaction} matrix, while also allows independent evaluation signals to enter through an \emph{external field} vector.
    Under the JC framework, we instantiate the interaction matrix using LLM-as-a-judge comparisons and consider alternative constructions, including Borda-JC. 

    \item \textbf{A unifying and scalable formulation.} 
    JC encompasses several existing TTA methods, including Self-Consistency and Weighted Self-Consistency, as special cases, while allowing independent and comparative evaluation signals to be combined.
    By varying the hyperparameter $\mu$, JC interpolates between independent-evaluation-only and interaction-only aggregation.
    This perspective identifies existing TTA methods as special cases within a broader family of aggregation rules, and enables systematic exploration of stronger alternatives by changing $\mu$.
    We further develop an efficient approximation strategy that retains the benefits of interaction modeling while making JC computationally scalable.

    \item \textbf{Consistent gains across math and code reasoning benchmarks.} 
    We evaluate JC on large-scale reasoning traces on math and code reasoning tasks, covering both heterogeneous crowdsourced settings such as MathArena and controlled homogeneous settings.
    Across extensive experiments, JC outperforms existing baselines across tasks, judge models, trace budgets, and trace-generation settings, with especially large gains on challenging problems and heterogeneous trace pools.
    Moreover, JC scales effectively with the number of candidate traces, remains robust to hyperparameter choices, and achieves the benefits of interaction modeling at costs comparable to WSC through efficient approximation.

\end{itemize}

\section{Prior Work and Preliminaries}
In this section, we formulate the test-time aggregation problem and review prior work most relevant to Joint Consistency. A more comprehensive discussion of related work is provided in Appendix~\ref{sec:related-work}.

\subsection{Problem Formulation of Test-Time Aggregation}

Consider a collection of $N$ reasoning traces $\{y_i\}_{i=1}^{N}$, where each trace $y_i = (z_i, a_i)$ consists of a \emph{reasoning chain} $z_i$ and an \emph{answer} $a_i$.
Let $\mathcal{A} = \{a^{(1)}, \dots, a^{(K)}\}$ denote the set of $K$ distinct candidate answers appearing in $\{a_i\}_{i=1}^N$.
The goal of Test-Time Aggregation (TTA) is to infer a single \emph{aggregated answer} $\hat{a} \in \mathcal{A}$.

We partition the set of traces according to their answers. Formally, let $\{\mathcal{I}_k\}_{k=1}^K$ be a partition of $[N] := \{1,\dots,N\}$ defined by
\begin{equation} \label{eq:Ik-def}
\mathcal{I}_k := \{\, i \in [N] \mid a_i = a^{(k)} \,\}.
\end{equation}
Let $\mathbf{x} \in \{0,1\}^{N}$ be an indicator vector, where $\mathbf{x}_i = 1$ indicates that the aggregated answer $\hat{a} = a_i$. 
Then, the feasible set of $\mathbf{x}$ is $\{ \mathbf{1}_{\mathcal{I}_k} \mid \ k \in [K] \}$, where $\mathbf{1}_{\mathcal{I}_k}$ denotes the indicator vector of the set $\mathcal{I}_k$.

\subsection{Evaluation-Free TTA} \label{sec:prior-free}
\paragraph{Self-Consistency (SC).} 
Self-Consistency~\citep{wang2023self,brown2024large,schaeffer2025large,zhou2025theoretical} is a classic \emph{evaluation-free} method that aggregates $\{y_i = (z_i, a_i) \}_{i=1}^{N}$ via majority vote over final answers $\{a_i\}_{i=1}^{N}$, treating all candidates with equal weight. 
It can be equivalently formulated as
\begin{align} \label{eq:SC}
    \min_{\mathbf{x} \in \{ \mathbf{1}_{\mathcal{I}_k} \mid \ k \in [K] \}} \quad - \langle \mathbf{h}, \mathbf{x} \rangle,
\end{align}
where $\mathbf{h} = \mathbf{1}_{N} \in \mathbb{R}^N$ is the all-one vector.
Let $\hat{\mathbf{x}}=\mathbf{1}_{\mathcal{I}_{\hat{k}}}$ be a minimizer of \eqref{eq:SC}; the aggregated answer is then $\hat{a}=a^{(\hat{k})}$.

\subsection{TTA with Independent Evaluation Signals} \label{sec:prior-ind}
In \emph{evaluation-based} TTA methods, evaluation is typically applied to the complete reasoning trace, using \emph{outcome reward models} (ORMs).  Evaluation signals may be \emph{intrinsic}, derived directly from the generator model, such as sequence log-probability, token-level certainty~\citep{kadavath2022language,kang2025scalable,taubenfeld2025confidence}, or verbalized confidence~\citep{tian2023just,xiong2024can}. 
However, such intrinsic evaluation signals are often not directly comparable across heterogeneous generators, and are usually unavailable in crowdsourced settings, limiting their applicability.
Alternatively, \emph{extrinsic} evaluation signals can be obtained from separate verifiers~\citep{cobbe2021training}, trained reward models~\citep{huang2025best,thatikonda2025logical,wan2025reasoning}, or from prompting-based evaluators such as LLM-as-a-judge \citep{zheng2023judging,gu2024survey,zhou2025evaluating,liu2025pairjudge}.

\paragraph{Weighted Self-Consistency (WSC).} 
Weighted Self-Consistency extends SC (\ref{eq:SC}) by encoding evaluation signals in the weight vector $\mathbf{h}$, where $\mathbf{h}_i$ is the weight of trace $y_i$. 

\paragraph{Self-Certainty.} Self-Certainty~\citep{kang2025scalable} aggregates traces based on the relative ranking of their confidence scores. However, these scores are computed \emph{independently} for each trace, without reference to other candidates. As a result, Self-Certainty does not leverage pairwise comparison signals and can be viewed as a special case of WSC, where $\mathbf{h}_i$ is given by a Borda-style rank-based score $\mathbf{h}_i = (N - r_i + 1)^q$, where $r_i$ denotes the rank of trace $y_i$ according to its self-certainty score, computed by averaging the token-level KL divergence from the uniform distribution to the model’s predictive distribution,
and $q > 0$ is a hyperparameter.

\subsection{TTA with Comparative Evaluation Signals} \label{sec:related_work_comparative}

The key distinction between \emph{independent} and \emph{comparative} evaluation signals lies in whether the judge model evaluates candidates \emph{jointly}. For instance, when employing LLM-as-a-judge, independent evaluation signals are elicited by assessing each trace in a separate query, whereas comparative evaluation signals are elicited by presenting two candidate traces together within one query context.

\textbf{Knockout Tournament (KT).} Knockout tournament~\citep{liu2025pairjudge} operates purely on comparative judgments. It performs aggregation through iterative pairwise comparisons, where pairwise judges determine relative correctness between candidates from distinct answer groups and progressively eliminate inferior ones until a single survivor remains. By avoiding exhaustive pairwise comparisons, this stochastic elimination procedure reduces the computational complexity to $O(N \log N)$.

\textbf{Global ranking from pairwise comparisons.}
Rather than eliminating candidates locally, standard rank-aggregation methods summarize a collection of pairwise judgments with a scalar score for each candidate. Copeland ranks candidates by their pairwise wins and losses~\citep{fishburn1977condorcet}, Elo updates candidate ratings sequentially according to comparison outcomes~\citep{elo1978rating}, and the Bradley--Terry model fits latent scores whose differences determine pairwise preference probabilities~\citep{bradley1952rank}; see Section \ref{sec:matharena-full}.

\section{Joint Consistency}\label{sec:joint_consistency}

We propose Joint Consistency (JC), a \emph{constrained energy minimization} approach for test-time aggregation inspired by the \emph{Ising} model:  
\begin{equation}\label{eq:ising_refined}
    \begin{aligned}
        & \min_{\mathbf{x} \in \{ \mathbf{1}_{\mathcal{I}_k} \mid \ k \in [K] \}} \quad H(\mathbf{x})  := - \mu \ \langle \mathbf{h}, \mathbf{x} \rangle -  \mathbf{x}^\top \mathbf{J} \mathbf{x},
    \end{aligned}
\end{equation}
where $\mathbf{h} \in\mathbb{R}^{N}$ encodes independent evaluation signals as described in Section \ref{sec:prior-ind},  $\mathbf{J}\in\mathbb{R}^{N\times N}$ encodes comparative evaluation signals, and $\mu>0$ controls the contribution of the former. 

The objective has an Ising-type structure, with $\mathbf{h}$ acting as
\emph{external fields} and $\mathbf{J}$ specifying pairwise
\emph{interactions}. Unlike the standard Ising model, the feasible set
restricts $\mathbf{x}$ to the indicator vector of a single answer group
$\mathcal{I}_k$, as defined in~(\ref{eq:Ik-def}). Figure~\ref{fig:Example-Ising} illustrates how JC leverages pairwise interactions to identify the correct answer when existing TTA methods fail.

The rest of this section is organized as follows.
Section~\ref{ssec:J} presents our construction of $\mathbf{J}$ based on pairwise comparative judgments, along with a soft Borda alternative in Section~\ref{sec:alternative_J}.
Section~\ref{ssec:approx} discusses the computational complexity of Joint Consistency and introduces an efficient approximation method.

\begin{figure}
    \centering
    \includegraphics[width=\linewidth,trim=1.0cm 8cm 12cm 8.8cm,clip,page=2]{Figures/diagram.pdf}
\caption{
TTA methods on crowdsourced traces for Q28 from the HMMT 2025 (Nov) dataset, where the ground-truth answer is \texttt{254}.
Yellow cells denote the final aggregated answer selected by each method. Blue and green cells represent independent evaluation signals and comparative signals, respectively.
The right table reports the answer-level scores used by each method:
SC counts supporting traces, BoN takes the highest independent score,
WSC sums independent scores, KT records the tournament round reached,
and JC computes the energy in \eqref{eq:ising_refined}.
Higher scores are preferred, except for JC, which minimizes energy.
SC, BoN, and WSC favor incorrect majority or high-scored answers.  KT eliminates the correct answer \texttt{254} in Round 2 after losing to \texttt{128} with a pairwise score of 0.4. In contrast, JC with $\mu=0.5$ successfully identifies the correct answer by aggregating global pairwise interactions. 
}
    \vspace{-1em}
    \label{fig:Example-Ising}
\end{figure}

\subsection{Interaction Modeling via LLM-as-a-Judge}\label{ssec:J}

We design the interaction matrix $\mathbf{J}$ as a positive semi-definite matrix that encodes comparative evaluation signals elicited from an LLM-as-a-judge. 

\textbf{Pairwise preference probabilities.}
Given two reasoning traces $y_i$ and $y_j$, the judge model is asked to estimate the probability that $y_i$ is better than $y_j$. 
Let $\pi_{\theta}$ denote the LLM-as-a-judge. We define
\begin{equation}\label{eq:ptheta-def}
    p_{\theta}(y_{i}\succeq y_j) := \mathbb{E}\left[ P_{\theta}(y_{i}\succeq y_j) \right]
\end{equation}
where $P_{\theta}(y_i \succeq y_j)$ denotes the normalized probability value output by $\pi_{\theta}$ of the event that $y_i$ is preferred to $y_j$ (see Appendix \ref{para:pair-order}), and the expectation is taken over the stochasticity of the judge model's responses. Concrete examples of the prompts used to query the judge model are provided in Appendix~\ref{sec:prompts}.

\paragraph{Positive semi-definite construction.}
We construct the interaction matrix as a Gram matrix:
\begin{equation}\label{eq:Jtheta}
    \mathbf{J}=\mathbf{C}\mathbf{C}^{\top},
\end{equation}
where $\mathbf{C}\in\mathbb{R}^{N\times N}$ has entries $C_{ij}:=\sqrt{p_\theta(y_i\succeq y_j)/(n_i^2n_j)}$ for all $i,j\in[N]$. Here, $n_i$ denotes the number of traces sharing the same final answer as $y_i$; that is, if $i\in\mathcal{I}_k$, then $n_i=|\mathcal{I}_k|$. This normalization balances answer-group sizes in $\mathbf{x}^{\top}\mathbf{J}\mathbf{x}$: the factor $n_i^2$ averages over trace pairs within the selected answer group, while $n_j$ averages over traces within each reference answer group.

Intuitively, each row of $\mathbf{C}$ can be viewed as a \emph{comparative profile} of a trace:
it represents how strongly the trace is preferred to each candidate.
Consequently, $\mathbf{J}_{ij} = \langle \mathbf{C}_{i,:}, \mathbf{C}_{j,:} \rangle$ measures the similarity between the comparative profiles of $y_i$ and $y_j$.
Thus, JC with \eqref{eq:Jtheta} tends to favor an answer group when its traces exhibit coherent pairwise preference patterns against the remaining candidates. 

The Gram construction admits an exact decomposition into a group-balanced soft Borda score and a within-group disagreement penalty, as shown in \eqref{eq:gram-borda-relation}. Appendix~\ref{sec:interpretation} provides a short derivation. We denote $P_{\theta}(y_i)$ as the independent evaluation score of trace $y_i$ elicited by LLM-as-a-judge $\pi_{\theta}$. The resulting Joint Consistency pipeline is summarized in Algorithm~\ref{alg:JC}.

\begin{algorithm}[H]
\caption{Joint Consistency for Test-Time Aggregation}
\label{alg:JC}
\begin{algorithmic}[1]
\STATE \textbf{Input:} Question $q$, number of traces $N$, judge model $\pi_\theta$,  $\mu$
\STATE Generate or collect $N$ reasoning traces $\{y_i=(z_i,a_i)\}_{i=1}^{N}$ for question $q$.
\STATE Partition traces by their extracted answers: 
$\{\mathcal{I}_k\}_{k=1}^{K}$, where $\mathcal{I}_k=\{i:a_i=a^{(k)}\}$.
\STATE Estimate $\mathbf{h}$ using independent trace-level evaluation scores $P_{\theta}(y_i)$ from $\pi_\theta$.
\STATE Estimate the interaction matrix $\mathbf{J}$ using pairwise comparative judgments $P_{\theta}(y_i \succeq y_j)$ from $\pi_\theta$.
\STATE $\hat{\mathbf{x}}=\mathbf{1}_{\mathcal{I}_{\hat{k}}} \leftarrow$  Solve the constrained Ising objective in \eqref{eq:ising_refined}
\STATE \textbf{Output:} Aggregated answer $\hat{a}=a^{(\hat{k})}$.
\end{algorithmic}
\end{algorithm}

\subsection{Alternative Constructions of the Interaction Matrix}\label{sec:alternative_J}
We now present a soft Borda alternative construction of the interaction matrix $\mathbf{J}$.

\paragraph{Soft Borda based interaction.}
Writing $p_{ij}:=p_\theta(y_i\succeq y_j)$, we construct a diagonal interaction matrix with
\begin{equation}\label{eq:Borda-J}
  (\mathbf{J}_{\mathrm{Borda}})_{ii} =
\displaystyle \sum_{j=1}^{N} \frac{p_{ij}}{n_i n_j}.  
\end{equation}
Under $\mu=0$, this recovers answer-level soft Borda scoring: for each answer $a^{(k)}$, the preference probabilities are averaged over trace pairs within each pair of answer groups, and these averages are then summed over the reference answer groups.

For the exact trace-level construction and $\mathbf{x}=\mathbf{1}_{\mathcal{I}_k}$,
\begin{equation}\label{eq:gram-borda-relation}
\mathbf{x}^{\top}\mathbf{J}\mathbf{x}
=\mathbf{x}^{\top}\mathbf{J}_{\mathrm{Borda}}\mathbf{x}
-\sum_{j=1}^{N}\frac{1}{n_j}
\operatorname{Var}_{i\sim\mathrm{Unif}(\mathcal{I}_k)}\!\left(\sqrt{p_{ij}}\right).
\end{equation}
Thus, Gram-JC subtracts a penalty for disagreement among traces supporting the same answer about a fixed reference trace. The penalty is zero when traces supporting the same answer have identical preferences against each reference trace. A derivation of \eqref{eq:gram-borda-relation} is provided in Appendix~\ref{sec:interpretation}. 
Further kernel-based constructions, including RBF-JC, are presented in Appendix~\ref{para:rbf-JC}.

\subsection{Computational Complexity and Efficient Approximation}\label{ssec:approx}
\textbf{Approximation strategy.}
A naive implementation of JC constructs $\mathbf{J}$ by evaluating pairwise comparisons over all trace pairs, incurring an $O(N^2)$ computational cost, which becomes prohibitive when the number of traces $N$ is large. 
To avoid this, we use an efficient approximation strategy: instead of exhaustively comparing all individual traces, we estimate interactions between answer groups by sampling only a constant number of traces from each group. The complexity is thus reduced from $O(N^2)$ to $O(K^2)$, where $K$ denotes the number of distinct candidate answers.

\textbf{Top-$\kappa$ approximation for scalability.}
For challenging problems, the number of distinct answers $K$ can still be large; see \Cref{tab:challenge_stats}. We can therefore further restrict pairwise evaluation to the top-$\kappa$ most frequent answers, where $\kappa$ acts as a \emph{consistency budget}. This reduces the pairwise comparison cost to $O(\kappa^2)$. Empirically, Figure~\ref{fig:pareto} and Table~\ref{tab:cost_table} in \Cref{ssec:challenge} show that a small fixed $\kappa$ already achieves strong performance. Answers outside the top-$\kappa$ set remain eligible for selection and can still be selected if their independent evaluation scores
are sufficiently high.

\textbf{Cost of LLM evaluation for one trace pair.}
LLM-as-a-judge evaluation of a trace pair places both traces in the input prompt while requiring only a small number of output tokens to produce the judgment. As input tokens are typically substantially cheaper than output tokens, pairwise evaluation is relatively cost-efficient. For example, on OpenRouter, output tokens are approximately $4.6\times$ more expensive when querying GPT-OSS-120B (\$0.18 versus \$0.039 per million).\footnote{\url{https://openrouter.ai/openai/gpt-oss-120b/pricing}}
In contrast, reasoning trace generation often incurs a much larger number of output tokens than input tokens. 

\textbf{Computational tractability for solving \eqref{eq:ising_refined}.}
In practice, the cost for solving \eqref{eq:ising_refined} is \emph{negligible} compared to the cost of generating reasoning traces and querying the judge model, since the feasible set is restricted to only $K$ configurations. With the top-$\kappa$ approximation, the complexity further reduces to $O(\kappa + N)$, where $N$ comes from the external term.

\section{Experiments}\label{sec:experiments}
We evaluate Joint Consistency on both math and code reasoning tasks. Our experiments cover two complementary settings. In the crowdsourced setting (Section~\ref{ssec:crowdsource}), we use
solutions from MathArena~\cite{balunovic2025matharena}, generated by diverse models for each problem. In the homogeneous setting (Section~\ref{ssec:homogeneous}), traces are generated by a single model, and token-level intrinsic evaluation signals, such as log-probabilities, are available.  This enables comparisons with intrinsic-evaluation-based TTA baselines.
Throughout the experiments, we use LLM-as-a-judge as an extrinsic evaluator for obtaining both independent and comparative evaluation signals. Unless otherwise stated, all experiments use the Gram construction of the interaction matrix $\mathbf{J}$ defined in \eqref{eq:Jtheta}; alternative constructions are explicitly labeled.
Detailed experimental settings, including prompt templates, are provided in Appendix~\ref{sec:prompts}.

\subsection{Crowdsourced Tasks}
\label{ssec:crowdsource}

We evaluate Joint Consistency on heterogeneous traces from MathArena, where candidate solutions to the same problem are generated by diverse models\footnote{
Traces are available at \url{https://huggingface.co/datasets/MathArena}. We use a revision where the four datasets are evaluated by 58, 38, 57, and 15 models, respectively. Each model provides four traces for every question.
}. Since these traces lack common intrinsic evaluation signals, such as token-level log-probabilities or process rewards, methods such as Self-Certainty, DeepConf, and PRM-based aggregation are not directly applicable.

We consider four competition-level math benchmarks, namely AIME'25~\citep{aops2025aimei,aops2025aimeii}, HMMT'25-Feb, HMMT'25-Nov, and Brumo'25. 
These benchmarks are challenging even for strong models; for instance, GPT-4o achieves only 14\% Pass@1 accuracy on HMMT'25-Feb setting, as shown in Fig.~\ref{fig:HMMT_Cost}. 

To evaluate performance across different levels of problem difficulty and generator quality, we construct three MathArena variants:
\begin{itemize}[leftmargin=2em]
    \item \textbf{MathArena-Full}: All problems and available model traces.
    \item \textbf{MathArena-C}:  A subset of challenging problems selected based on low Pass@1 accuracy and high answer diversity, measured by the average number of traces per distinct answer. Dataset statistics are provided in Table~\ref{tab:challenge_stats}.
    \item \textbf{MathArena-W}: Traces generated by the lowest-performing models in MathArena.
\end{itemize}

\subsubsection{Scalable \texorpdfstring{$\kappa$}{kappa}-Approximation of the Interaction Matrix \texorpdfstring{$\mathbf{J}$}{J}}\label{ssec:challenge}

\begin{wrapfigure}{r}{0.5\textwidth}
    \centering
    \vspace{-2em}
    \includegraphics[width=\linewidth]{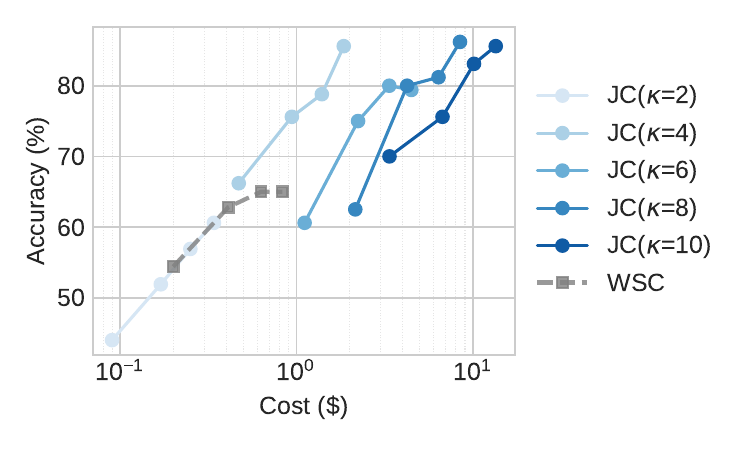}
    \caption{\textbf{Accuracy--cost trade-off under $\kappa$-approximation on MathArena-C.}
 Only evaluation costs are included.}
     \vspace{-2em}
    \label{fig:pareto}
\end{wrapfigure}

We investigate how the consistency budget $\kappa$ affects the cost and accuracy of Joint Consistency on MathArena-C. Specifically, we vary the consistency budget $\kappa \in \{2,4,6,8,10\}$ and use $\{20\%,40\%,60\%,80\%\}$ of the available reasoning traces for test-time aggregation. We set $\mathbf{h}\equiv 0$ to isolate the effect of the interaction matrix $\mathbf{J}$, and use GPT-OSS-20B as the judge for both JC and WSC. For a fair cost comparison, we report the actual API usage for both independent and comparative evaluations.

The results are shown in \Cref{fig:pareto}. We observe that \textbf{a small $\kappa$ is sufficient for JC to scale beyond WSC.} JC surpasses the performance plateau of WSC and achieves an accuracy gain of over 20\%, even with a small budget of $\kappa=4$. Moreover, across different values of $\kappa$, JC continues to improve as the evaluation budget increases, suggesting that a small subset of frequent answers is sufficient to capture most of the benefit of JC. Moreover, the results show that \textbf{the interaction $\mathbf{J}$ can be approximated efficiently}. Detailed results are provided in Table~\ref{tab:cost_table} in the appendix.


\begin{figure}[t]
    \centering
    \begin{minipage}[t]{0.48\textwidth}
        \centering
        \includegraphics[width=\linewidth]{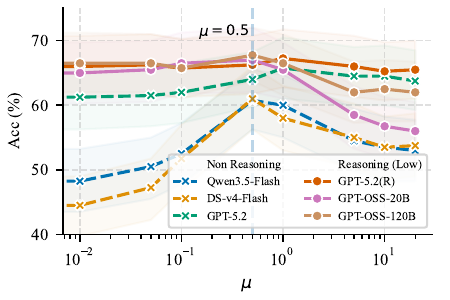}
        \caption{ 
        Sensitivity of Joint Consistency to $\mu$ on MathArena-C dataset. Across diverse judge models, JC performs best with a moderate $\mu \in [0.5,1]$, suggesting that \textbf{$\mu$ does not require judge-model-specific tuning}.
        }
            \label{fig:mu}
    \end{minipage}
    \hfill
    \begin{minipage}[t]{0.48\textwidth}
        \centering
        \includegraphics[width=\linewidth]{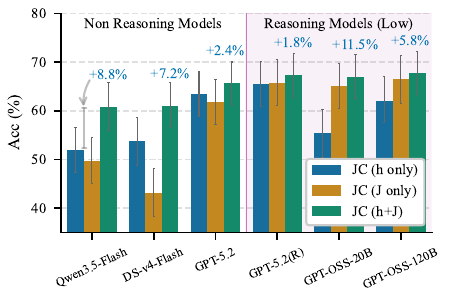}
        \caption{Accuracy of Joint Consistency across different judge models on MathArena-C dataset. Bars compare three variants of JC, namely $\mathbf{h}$ only, $\mathbf{J}$ only, and $\mathbf{h}$+$\mathbf{J}$, where the $\mathbf{h}$ only variant is WSC. The annotated values ({\color{RoyalBlue!80}+X\%}) show the accuracy gain of JC ($\mathbf{h}$+$\mathbf{J}$) over WSC. 
        }
        \label{fig:bar}
    \end{minipage}
\end{figure}

\subsubsection{Robustness to Diverse Judge Models}

We evaluate Joint Consistency with diverse judge models on MathArena-C. 
We consider five judge model families: Qwen3.5-Flash~\citep{qwen3.5}, DeepSeek-V4-Flash~\citep{deepseekai2026deepseekv4}, GPT-5.2~\citep{openai2025gpt52}, GPT-OSS-20B, and GPT-OSS-120B~\citep{openai2025gptoss120bgptoss20bmodel}. 
For Qwen, DeepSeek, and the default GPT-5.2 setting, we disable reasoning. 
For GPT-OSS-20B and GPT-OSS-120B, we use low reasoning effort. 
We also include GPT-5.2 with low reasoning effort, denoted as GPT-5.2(R).

\textbf{The optimal $\mu$ is stable across judge models.}
For each judge model, we evaluate Joint Consistency over 
$\mu \in \{0, 0.01, 0.05, 0.1, 0.5, 1, 5, 10, 20\}$.
As shown in \Cref{fig:mu}, the best performance is generally achieved between $0.5$ and $1$, across different judges.
This suggests that $\mu$ does not require careful tuning for a specific judge model.

Although GPT-5.2 with low reasoning effort achieves the strongest overall performance across most values of $\mu$, it is $19\times$ more expensive than GPT-OSS-120B in our setting and provides only modest additional gains.
This indicates that frontier closed-source models are not necessary for effective judging: open-weight GPT-OSS models offer competitive performance at substantially lower cost. 

\textbf{JC improves over WSC and its ablations for every judge model.}
As shown in \Cref{fig:bar}, JC \emph{consistently} outperforms the $\mathbf{h}$-only (WSC) and $\mathbf{J}$-only variants across all judge models, and improves over WSC by $+1.8\%$ to $+11.5\%$. While reasoning improves pairwise evaluation and makes the $\mathbf{J}$-only variant more competitive, combining $\mathbf{h}$ and $\mathbf{J}$ remains the most reliable choice, confirming that independent and comparative evaluation signals provide complementary information for TTA. 

\subsubsection{Aggregation over Complete Heterogeneous Trace Pools} \label{sec:matharena-full}

We evaluate JC on MathArena-Full, which contains all available problems and model-generated traces. We compare JC and its Borda-JC variant against eight baselines: Pass@1, Best-of-N (BoN), Self-Consistency (SC), Weighted Self-Consistency (WSC), Knockout Tournament (KT), Copeland, Elo, and Bradley--Terry (BT). The three global ranking baselines use the same pairwise judgments as JC and require no additional judge calls. JC and Borda-JC are evaluated on identical trace samples and random seeds. Implementation details are provided in Appendix~\ref{app:ranking_baselines}.

JC and Borda-JC achieve comparable overall performance, with neither construction consistently outperforming the other across datasets and sampling ratios. At least one of the two achieves the best or tied-best performance in all 12 settings of Table~\ref{tab:matharena_table}. Moreover, the additional pairwise evaluation cost is significantly less than the trace-generation cost.

\begin{table}[H]
    \centering
    \caption{\textbf{Accuracies of Joint Consistency on MathArena (Full)}.
    The GPT-OSS-20B model with low reasoning effort is used to score both $\textbf{h}$ and $\textbf{J}$. Here, $\rho$ denotes the fraction sampled from each dataset's trace pool.}
    \footnotesize
    \begin{adjustbox}{max width=\textwidth}
     {\input{Tables/5-1_main}}
    \end{adjustbox}
    \label{tab:matharena_table}
\end{table}

\paragraph{Performance on harder problems.}
\begin{wraptable}{r}{0.48\textwidth}
    \centering
    \footnotesize
    \setlength{\tabcolsep}{3pt}
    \caption{\textbf{Accuracy on the hardest MathArena-Full difficulty bin.} Results average over 20 questions (five per benchmark) and the same ten seeds as Table~\ref{tab:matharena_table}. $\Delta_{\mathrm{WSC}}$ and $\Delta_{\mathrm{Borda}}$ are JC's gains in percentage points. Bold marks the highest accuracy in each row.}
    \begin{adjustbox}{max width=\linewidth}
    \begin{tabular}{crrrrr}
    \toprule
    $\rho$ & WSC & Borda-JC & JC & $\Delta_{\mathrm{WSC}}$ & $\Delta_{\mathrm{Borda}}$ \\
    \midrule
    $10\%$ & $78.0$ & $\mathbf{81.5}$ & $80.5$ & $+2.5$ & $-1.0$ \\
    $20\%$ & $80.0$ & $82.0$ & $\mathbf{84.0}$ & $+4.0$ & $+2.0$ \\
    $50\%$ & $81.5$ & $82.0$ & $\mathbf{84.5}$ & $+3.0$ & $+2.5$ \\
    \bottomrule
    \end{tabular}
    \end{adjustbox}
    \label{tab:hardest-bin-main}
\end{wraptable}
To examine where aggregation gains arise, we divide each benchmark into six difficulty bins ranked by full-pool Pass@1, used only for stratification. Table~\ref{tab:hardest-bin-main} summarizes the hardest bin, containing the five hardest questions from each benchmark. JC improves over WSC by $2.5$--$4.0$ percentage points across sampling ratios. Compared with Borda-JC, it is one point lower at $\rho=10\%$ but $2.0$ and $2.5$ points higher at the two larger ratios. The other five bins are largely saturated, with JC and WSC differing by at most one point. These results suggest that the benefit is concentrated on harder problems in this trace pool. Full results are provided in Appendix~\ref{ssec:difficulty-analysis}.

\subsubsection{Robustness to Lower-Tier Trace Generators}
We investigate whether JC remains effective when reasoning traces are generated by lower-performing models. We rank models by their Pass@1 accuracy on AIME 2025, retain the $R$ lowest-performing models, and evaluate $R\in\{3,5,10,15,20,30\}$.

\textbf{JC remains effective with lower-tier trace generators.}
As shown in Table~\ref{tab:worst}, JC and Borda-JC outperform the competing baselines on average, both achieving $61.89\%$, compared with $56.91\%$ for the strongest non-JC baseline, WSC. Borda-JC performs best at $R=20$ and $R=30$. These results suggest that JC can effectively aggregate reasoning traces even when they are generated by relatively weak models.


\begin{table}[t]
    \centering
    \caption{\textbf{Accuracies of TTA methods on traces from lower-tier models on AIME 2025.} $R$ denotes the $R$ lowest-ranked AIME 2025 generators. We use 50\%-per-problem sampling. Copeland, Elo, soft-outcome Bradley--Terry (BT), Borda-JC and JC use the complete answer-pair comparison graph; KT uses pairwise tournament elimination. Entries report mean accuracies, and the final row is the average over the six values of $R$; the corresponding standard-deviation statistics are reported in Appendix Table~\ref{tab:worst_full}.}
    \input{Tables/5-2_main}

    \label{tab:worst}
\end{table}

\subsection{Comparing Intrinsic-Evaluation TTA Baselines in Homogeneous Settings}
\label{ssec:homogeneous}

We now consider homogeneous settings where traces are generated together with token-level log-probabilities by the same model, making intrinsic-evaluation-based TTA methods applicable. 
We thus add two representative baselines: Self-Certainty~\cite{kang2025scalable} implemented with Borda-$p$ voting using $p=2$, and DeepConf~\cite{fu2025deep} implemented with tail confidence.
We evaluate this setting on math reasoning tasks and code reasoning tasks. 


For math reasoning, we conduct experiments on AIME 2025.
We use Qwen3-8B, Qwen3-32B~\citep{yang2025qwen3}, and DeepSeek-R1-8B~\citep{guo2025deepseek} as trace generators, and DeepSeek-R1-8B and GPT-OSS-20B~\citep{openai2025gptoss120bgptoss20bmodel} as judge models. For each problem, we generate an initial pool of 256 reasoning traces and uniformly subsample 64 traces for aggregation. Detailed hyperparameters are provided in Appendix~\ref{sec:hyperparameters}.
As shown in Table~\ref{tab:cross-gen-eva}, JC outperforms all other baselines across all trace generators and judge models. 
 

For code reasoning problems, we evaluate on CruxEval-O~\citep{gu2024cruxeval}, a benchmark in which the model is given a Python function and an input and must predict the corresponding program output. The dataset contains 800 questions. We use Qwen3-Coder-30B to generate 32 traces for each question and resample $N \in \{4,8,16,24\}$ traces for test-time scaling. To make comparison fair, we constrain the computational budget of JC to scale linearly with $N$. The results are shown in Table \ref{tab:cruxeval_o}, from which we observe that JC outperforms all baselines for every $N$. Moreover, the gain $\Delta$ over the strongest baseline increases with $N$, indicating that JC benefits more effectively from larger test-time budgets.

The above results show that \textbf{JC remains effective even when competing baselines have access to intrinsic evaluation signals, and continues to scale favorably as the number of traces increases.}

\begin{table}[t]
\centering

\begin{minipage}[t]{0.48\textwidth}
\centering
\caption{\textbf{TTA accuracy on AIME 2025.}
JC outperforms intrinsic-evaluation baselines across trace generators and judge models.}
\label{tab:cross-gen-eva}
\resizebox{\linewidth}{!}{%
\begin{tabular}{ccccc}
\toprule
&& \multicolumn{3}{c}{{Trace Generation Models}}\\
\cmidrule(lr){3-5}
\multirow{2}{*}{\makecell{Judge \\Models}} 
& \multirow{2}{*}{\makecell{TTA\\Methods}} 
& \multicolumn{2}{c}{\textbf{Qwen3}} 
& \textbf{DS-R1} \\
& & \textbf{8B} & \textbf{32B} & \textbf{8B} \\
\cmidrule(lr){1-5}
\multirow{3}{*}{-} 
& SC         & $77.0$ & $80.2$ & $83.1$ \\
& Self-Cert. & $76.6$ & $80.0$ & $83.1$ \\
& DeepConf   & $76.8$ & $80.3$ & $83.1$ \\
\cmidrule(lr){1-5}
\multirow{2}{*}{\makecell{DS-R1\\8B}}
& WSC        & $76.8$ & $80.7$ & $85.5$ \\
& JC         & $\mathbf{84.1}$ & $\mathbf{84.9}$ & $\mathbf{87.1}$ \\
\cmidrule(lr){1-5}
\multirow{2}{*}{\makecell{GPT-OSS\\20B}}
& WSC        & $81.4$ & $83.4$ & $88.4$ \\
& JC         & $\mathbf{86.2}$ & $\mathbf{90.9}$ & $\mathbf{93.5}$ \\
\bottomrule
\end{tabular}
}
\end{minipage}
\hfill
\begin{minipage}[t]{0.49\textwidth}
\centering
\small
\caption{\textbf{TTA accuracy on CruxEval-O code reasoning.}
$\Delta$ denotes JC's gain over the strongest non-JC baseline.}
\label{tab:cruxeval_o}
\resizebox{\linewidth}{!}{%
\begin{tabular}{c|cccc}
\toprule
\textbf{Method} 
& \textbf{$N$=$4$} 
& \textbf{$N$=$8$} 
& \textbf{$N$=$16$} 
& \textbf{$N$=$24$} \\
\midrule
Pass@1 & 71.3 & 71.3 & 71.3 & 71.3 \\
BoN    & 76.2 & 77.0 & 78.0 & 78.0 \\
SC     & 74.7 & 76.0 & 76.3 & 76.4 \\
WSC    & \underline{76.9} & \underline{77.8} & \underline{78.1} & \underline{78.1} \\
Self-Cert. & 73.5 & 74.7 & 75.6 & 76.0 \\
DeepConf  & 71.9 & 74.7 & 75.6 & 75.8 \\
\midrule
JC     & \textbf{77.3} & \textbf{78.7} & \textbf{79.9} & \textbf{80.5} \\
$\Delta$ & +0.4 & +0.9 & +1.8 & +2.4 \\
\bottomrule
\end{tabular}
}
\end{minipage}

\end{table}
\section{Conclusion}\label{sec:conclusion}
We introduced Joint Consistency, a unified test-time aggregation framework that integrates independent evaluations with pairwise comparative signals under a constrained Ising formulation. JC highlights interaction modeling as an important component that is largely absent from existing TTA methods. Our experiments showed that JC outperforms strong baselines in both crowdsourced and homogeneous settings, while incurring only marginal additional cost via our efficient approximation strategy. 

\paragraph{Limitations.} 
This work has several limitations. First, our paper focuses on final-answer aggregation in reasoning tasks, and it remains unclear to what extent the observed improvements generalize to open-ended reasoning or research-level problem solving. Second, although we provide a theoretically motivated construction of the interaction matrix $\mathbf{J}$ in Section~\ref{ssec:J}, this design is not necessarily optimal. Exploring alternative constructions of $\mathbf{J}$ remains an important direction for future work. Finally, our method may inherit the limitations and potential risks of the underlying generation and judge models.

\begin{ack}
Michael Gastpar was supported by the Swiss National Science Foundation (SNSF) under Grants 236494 and 10004381. Bo Jiang was supported in part by the National Natural Science Foundation of China (NSFC) under Grants 72394364 and 72394360. Lie He was supported in part by the National Natural Science Foundation of China (NSFC) under Grant 62606302 and by IQuest Research.
\end{ack}
\bibliography{ref}
\bibliographystyle{plain}


\newpage
\appendix

\section{Related Work} \label{sec:related-work}

Test-time scaling (TTS) improves accuracy on complex reasoning tasks for LLMs by allocating additional compute for sampling and aggregating candidate solutions into a final decision \citep{snell2024scaling,welleck2024decoding,balachandran2025inference}. Existing TTS methods can be categorized along three aspects. First, depending on whether evaluation is employed, a TTS method can be \emph{evaluation-free} or \emph{evaluation-based}. Second, evaluation can be applied to the complete reasoning trace, known as \emph{outcome reward models} (ORMs), or to intermediate reasoning steps, corresponding to \emph{process reward models} (PRMs). Finally, TTA can be implemented \emph{in parallel}, where candidate solutions are generated independently and aggregated post hoc, or \emph{sequentially}, where generation, evaluation, and refinement are interleaved. 

Self-Consistency (SC)~\citep{wang2023self,brown2024large,schaeffer2025large,zhou2025theoretical} is a classic \emph{evaluation-free} method that aggregates multiple sampled reasoning paths via majority voting over final answers, treating all candidates with equal weight. Subsequent work extends SC by incorporating more structured aggregation mechanisms. Semantic self-consistency~\citep{knappe2024semantic} replaces string-level voting with semantic-aware clustering and aggregation while universal self-consistency~\citep{chen2024universal} generalizes SC to free-form generation by prompting the LLM to select the final answer most consistent with the overall set.

Besides the most relevant comparative evaluation-based methods discussed in Section \ref{sec:related_work_comparative}, some papers employ listwise and adaptive comparison.
GenSelect~\citep{toshniwal2025genselect} performs a listwise comparison by placing multiple candidate traces in a shared context and directly selecting the best candidate. V$_1$-Infer~\citep{singhv1} instead uses pairwise comparisons adaptively: it first collects enough comparisons to estimate initial candidate scores and then compares similarly ranked candidates in a Swiss-tournament stage.

PRM-based methods incorporate state- or step-level evaluation directly into the generation process, actively guiding reasoning via inference-time search algorithms. In Tree of Thoughts (ToT)~\citep{yao2023tree, long2023large}, intermediate reasoning states are evaluated either using task-specific heuristics or by prompting the LLM itself to assess the path promise. Monte Carlo Tree Search (MCTS) based methods~\citep{hao2023reasoning,wan2024alphazero} use step-level value estimation to prune unpromising branches and navigate the reasoning space. More recent work explores confidence-aware generation strategies, where intrinsic uncertainty estimates are used to dynamically control reasoning depth and trajectories during inference~\citep{zhang2025confidence,fu2025deep}. 
PRM-based methods rely on generation-time signals and thus fall outside the scope of the post-hoc aggregation setting, especially in crowdsourced scenarios where such information is typically unavailable. 

In contrast to PRM-based methods that assess intermediate reasoning steps, ORM-based methods weight or rank candidate solutions using \emph{solution-level evaluation} signals. These signals may be \emph{intrinsic}, derived directly from the generator model, such as sequence log-probability, token-level certainty~\citep{kadavath2022language,kang2025scalable,taubenfeld2025confidence}, or verbalized confidence~\citep{tian2023just,xiong2024can}. 
However, such intrinsic evaluation signals are often not directly comparable across heterogeneous generators, and are usually unavailable in crowdsourced settings, limiting their applicability.
Alternatively, \emph{extrinsic} evaluation signals can be obtained from separate verifiers~\citep{cobbe2021training}, trained reward models~\citep{huang2025best,thatikonda2025logical,wan2025reasoning}, prompting-based evaluators such as LLM-as-a-judge \citep{zheng2023judging,gu2024survey,zhou2025evaluating,liu2025pairjudge}. Beyond fixed-budget parallel aggregation, some ORM-based methods adopt adaptive sampling strategies, where evaluation statistics computed over previously sampled candidates are used to dynamically determine when to stop further sampling, thus improving inference efficiency~\cite{shi2025reasoning}.

Sequential ORM-based approaches iteratively verify, critique, and refine candidate solutions using evaluation feedback, progressively steering the reasoning process toward correctness~\citep{chen2025sets}. 
Representative methods include self-verification~\citep{madaan2023self,weng2023large,ferraz2024llm,zhao2025sample} and self-correction~\citep{qu2024recursive,gou2024critic,lee2025evolving}.

\section{Hyperparameter Settings}\label{sec:hyperparameters}

The experiments were conducted on a server with an AMD EPYC 7402 24-Core processor, 256\,GB of memory, and four NVIDIA RTX A6000 GPUs, running Ubuntu 22.04.5 LTS and Python 3.12.11.

\subsection{Hyperparameters of Trace Generation Models}
Table~\ref{tab:generation_hyperparams} summarizes the decoding hyperparameters used during generation for all models. For each model, the temperature, and maximum generation length are fixed across experiments, and we use the model’s native tokenizer. A dash (—) denotes that the corresponding decoding option is disabled. The Reasoning budget (API param) column specifies how each model’s reasoning behavior is controlled during generation. For models that expose explicit API-level controls, we report the corresponding parameter setting (e.g., \texttt{reasoning\_effort} for GPT-OSS). For models without a dedicated reasoning parameter, reasoning is enabled either by default or via prompt-level instructions.

\begin{table}[H]
\centering
\small
\caption{Hyperparameters and statistics of trace generation models.}
\label{tab:generation_hyperparams}
\resizebox{\textwidth}{!}{
\begin{tabular}{llccc}
\toprule
Model & Task &Temperature & Max seq len & Reasoning budget (API param) \\
\midrule
Qwen3-8B   & Tab.~\ref{tab:cross-gen-eva}  & 0.6  & 40k & True (\texttt{enable\_thinking})  \\
Qwen3-32B  & Tab.~\ref{tab:cross-gen-eva}     & 0.6  & 40k & True (\texttt{enable\_thinking}) \\
DeepSeek-8B &  Tab.~\ref{tab:cross-gen-eva}  & 0.6 & 130k & ---\\
Qwen3-Coder-30B & Tab.~\ref{tab:cruxeval_o} & 0.7 &  256k & ---\\
\bottomrule
\end{tabular}
}
\end{table}

\subsection{Hyperparameters of Judge Models}
Table~\ref{tab:scoring_hyperparams} summarizes the decoding hyperparameters used during scoring for all models. For each model, the temperature, and maximum generation length are fixed across experiments, and we use the model’s native tokenizer. The Reasoning budget (API param) column specifies how each model’s reasoning behavior is controlled during generation. For models that expose explicit API-level controls, we report the corresponding parameter setting (e.g., \texttt{reasoning\_effort} for GPT-OSS). For models without a dedicated reasoning parameter, reasoning is enabled either by default or via prompt-level instructions.

\begin{table}[!ht]
\centering
\caption{Hyperparameters and statistics of judge models.}
\label{tab:scoring_hyperparams}
\resizebox{\textwidth}{!}{
\begin{tabular}{llccc}
\toprule
Model & Task& Temperature  & Max seq len & Reasoning budget (API param) \\
\midrule
DeepSeek-8B & Tab.~\ref{tab:cross-gen-eva} & 0.7 &  40k & --- \\
GPT-OSS-20B & Tab.~\ref{tab:matharena_table},~\ref{tab:worst},~\ref{tab:cross-gen-eva},~\ref{tab:cruxeval_o},~\ref{tab:cost_table}, Fig. \ref{fig:pareto},~\ref{fig:mu},~\ref{fig:bar}  & 1.0 & 128k & low (\texttt{reasoning\_effort})\\
GPT-OSS-120B &  Fig. \ref{fig:pareto},~\ref{fig:mu},~\ref{fig:bar} & 0.7 & 128k & low (\texttt{reasoning\_effort})\\
GPT-5.2  & Fig. \ref{fig:mu},~\ref{fig:bar} & 0.7 & 400k & w/, w/o low (\texttt{reasoning\_effort})\\
Qwen3.5-Flash  & Fig. \ref{fig:mu},~\ref{fig:bar} & 0.7 & 1000k & w/o\\
DeepSeek-V4-Flash  & Fig. \ref{fig:mu},~\ref{fig:bar} & 0.7 & 1000k & w/o\\
\bottomrule
\end{tabular}
}
\end{table}

\subsection{Hyperparameters and Settings of Aggregation Methods}
\label{app:ranking_baselines}

Tables~\ref{tab:matharena_table} and~\ref{tab:worst} include three global ranking baselines: Copeland, Binary Elo, and Bradley--Terry (BT). In Table~\ref{tab:matharena_table}, they use the top-$4$ answer filter; in Table~\ref{tab:worst}, they use the complete answer-pair comparison graph. In both cases, they reuse the pairwise judgments available to JC, operate on candidate answers, and require no additional judge calls.

\paragraph{Copeland ranking.}
For each pair of candidate answers, we average the available pairwise preference probabilities in each direction. A candidate receives a win if its mean preference exceeds that of its opponent, a loss if it is lower, and a tie otherwise. The Copeland score of candidate $i$ is $C_i = W_i-L_i$, where $W_i$ and $L_i$ denote its numbers of wins and losses against distinct opponents. Candidates are ranked by their Copeland scores.
This method is order-invariant but discards preference magnitudes.

\paragraph{Binary Elo.}
We initialize each candidate's rating to $1000$ and convert pairwise preference probabilities into binary outcomes: $S_i=1$ for a win, $S_i=0$ for a loss, and $S_i=0.5$ for a tie. We process one deterministically shuffled pass over all comparisons. For candidates $i$ and $j$, the expected outcome is
\[
E_i = \frac{1}{1+10^{(R_j-R_i)/400}},
\]
and the ratings are updated as
\[
R_i \leftarrow R_i + 400(S_i-E_i), \qquad
R_j \leftarrow R_j - 400(S_i-E_i).
\]
Candidates are ranked by their final ratings. Unlike Copeland, Elo depends on the order of comparisons, and its binary outcomes discard preference magnitudes.

\paragraph{Bradley--Terry ranking.}
We assign each candidate $i$ a latent strength $\theta_i$ and model its pairwise preference probability as
\[
P(i\succ j)=\sigma(\theta_i-\theta_j),
\]
where $\sigma$ is the logistic function. We use the same bidirectionally normalized soft preference $q_{ij}\in[0,1]$ as JC and estimate the latent strengths by minimizing the fractional binary cross-entropy
\[
-\sum_{(i,j)}\left[q_{ij}\log\sigma(\theta_i-\theta_j)+(1-q_{ij})\log\bigl(1-\sigma(\theta_i-\theta_j)\bigr)\right]
+\frac{\lambda}{2}\|\theta\|_2^2,
\]
where $\lambda=0.01$. We use BFGS for optimization and rank candidates by their fitted latent strengths. Unlike Copeland and Binary Elo, BT retains preference magnitudes and is independent of comparison order.

\section{Dataset Statistics}
\subsection{Pass@1 Accuracy of LLM-as-a-Judge}
\begin{figure*}[h]
    \centering
    \includegraphics[width=0.95\linewidth]{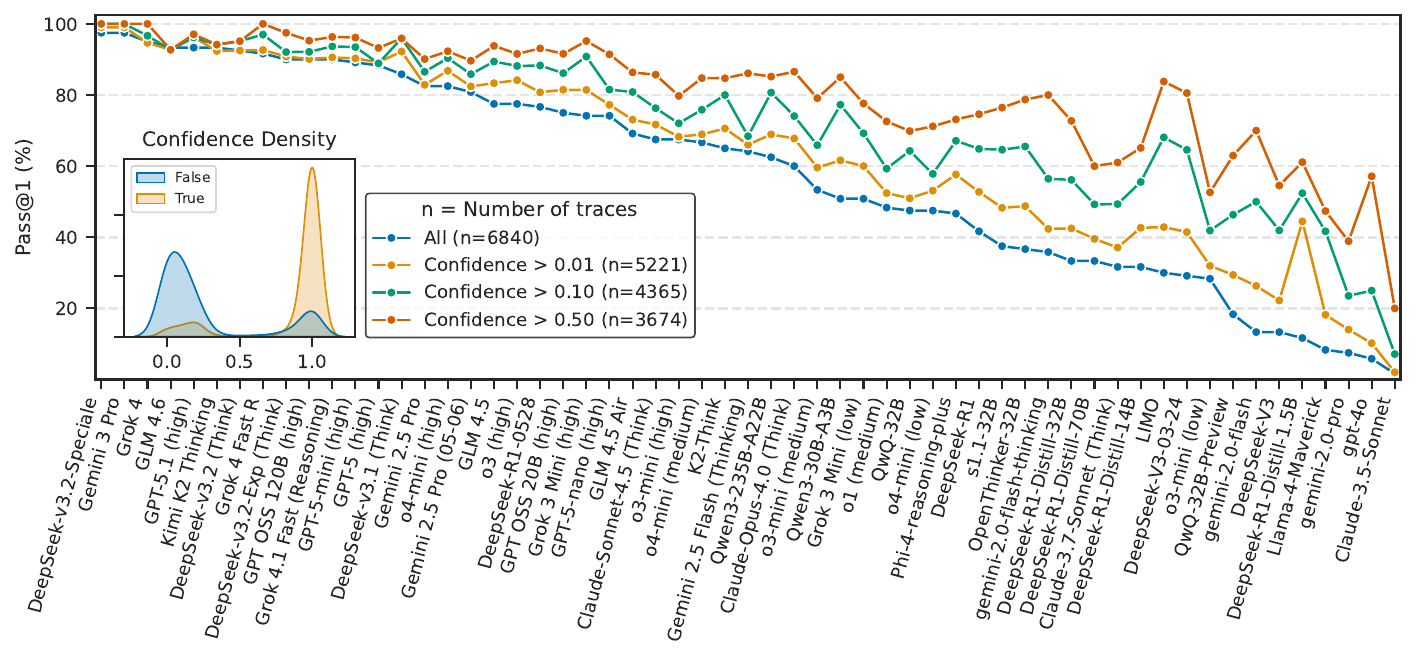}
    \caption{
    Pass@1 accuracies of reasoning traces from 57 models on the HMMT-2025 (Feb). 
    Confidence scores for these traces are evaluated using a GPT-OSS-20B model with low reasoning effort. The inset plot illustrates the score density functions for correct versus incorrect reasoning traces across the full set of 6,840 traces. 
    }
    \label{fig:HMMT_Cost}
\end{figure*}

Figure \ref{fig:HMMT_Cost} shows Pass@1 accuracies of reasoning traces from 57 models on the HMMT-2025 (Feb), measuring the effectiveness of different judge models by scoring traces. 

\subsection{Statistics of MathArena-C}
Questions used in MathArena-C are detailed in Table~\ref{tab:challenge_stats}. The Pass@1 accuracy is calculated across all available models in MathArena.
\begin{table}[H]
    \centering
    \small
    \caption{Statistics for the challenging subset of questions. Higher values ($\uparrow$) for \#Traces/\#Ans and Pass@1 typically indicate \emph{lower} problem difficulty. }
    
    {
    \begin{tabular}{llrr}
        \toprule
        Dataset & Q Idx &  \#Traces/\#Ans ($\uparrow$) & Pass@1 ($\uparrow$) \\
        \midrule
        \multirow[t]{3}{*}{AIME} & 14 & 228/75$\approx$3.0 & 13.6\% \\
         & 15 & 228/78$\approx$2.9 & 3.5\% \\
         & All& Avg$\approx$7.0& 69.8\% \\
        \midrule
        \multirow[t]{3}{*}{BRUMO} & 13 & 152/20$\approx$7.6 & 46.1\% \\
         & 30 & 152/48$\approx$3.2 & 21.1\% \\
         & All& Avg$\approx$14.7 & 86.1\% \\
        \midrule
        \multirow[t]{3}{*}{HMMT Feb} & 19 & 228/54$\approx$4.2 & 10.5\% \\
         & 20 & 228/103$\approx$2.2 & 12.7\% \\
         & All& Avg$\approx$5.8 & 60.0\% \\
        \midrule
        \multirow[t]{3}{*}{HMMT Nov} & 10 & 60/11$\approx$5.5 & 6.7\% \\
         & 28 & 60/20=3.0 & 11.7\% \\
         & All& Avg$\approx$14.4 & 87.8\% \\
        \bottomrule
    \end{tabular}
    }
    \label{tab:challenge_stats}
\end{table}

\section{Additional Experimental Results}

\subsection{Cost and Accuracy with Interaction Matrix Simplification}

Table~\ref{tab:cost_table} lists the data used to generate Figure \ref{fig:pareto}.
\begin{table*}[h]
    \centering
    \caption{\textbf{Accuracy (\%) and cost (USD) across varying consistency budgets ($\kappa$) and trace subsample ratio ($N$) on challenging crowdsourced problems.} Each cell displays the format ``Accuracy(\%)/\$Cost''. {\color{red} Red} text denotes the highest accuracy achieved for a specific $N$. Cells highlighted with a \colorbox{green!20}{green} background indicate configurations that outperform the best overall WSC accuracy ({\color{blue}65\%}). Note that while WSC is theoretically independent of $\kappa$, minor fluctuations in its cost and performance are due to the sampling stochasticity of the traces. 
    }
    \resizebox{\textwidth}{!}
    {%

\input{Tables/cost_table.tex}
    }
    \label{tab:cost_table}
\end{table*}

We also examine the effect of $\kappa$ when correct answers are infrequent, using traces from lower-tier generators.
Table \ref{tab:answer_coverage} lists the number of questions out of 30 questions such that the number of traces of the correct answer is within top-$\kappa$ frequent traces generated by the $R$ low-tier models in column 2-4. The last column shows the number of questions for which the correct answer appears among the traces generated by the $R$ models. 

\begin{table}[H]
\centering
\small
\caption{Number of questions (out of 30) for which the correct
answer appears among the top-$\kappa$ most frequent answers
in traces generated by the $R$ low-tier models.
``All'' considers all generated answers.}
\label{tab:answer_coverage}
\begin{tabular}{rcccc}
\toprule
$R$ & Top-$1$ & Top-$4$ & Top-$8$ & All \\
\midrule
 3 &  7 &  8 & 10 & 11 \\
 5 &  9 & 12 & 12 & 13 \\
10 & 13 & 17 & 22 & 22 \\
15 & 17 & 23 & 25 & 27 \\
20 & 24 & 25 & 25 & 28 \\
30 & 25 & 27 & 29 & 29 \\
\bottomrule
\end{tabular}
\end{table}

Table~\ref{tab:kappa_low_tier} compares WSC with JC on traces from the $R$ lowest-ranked generators. 
\begin{table}[H]
\centering
\small
\caption{Accuracy comparisons across values of $R$ and consistency budgets $\kappa$.}
\label{tab:kappa_low_tier}
\begin{tabular}{rcccc}
\toprule
$R$ & WSC & $\kappa=4$ & $\kappa=8$ & $\kappa=K$ \\
\midrule
 3 & 23.5 & 23.5 & 25.7 & 27.0 \\
 5 & 31.8 & 36.0 & 36.3 & 36.3 \\
10 & 51.6 & 50.7 & 61.3 & 59.7 \\
15 & 68.7 & 73.0 & 79.0 & 76.0 \\
20 & 77.7 & 82.7 & 83.0 & 82.3 \\
30 & 88.2 & 89.7 & 94.0 & 90.0 \\
\bottomrule
\end{tabular}
\end{table}



\subsection{Performance across Problem Difficulty}\label{ssec:difficulty-analysis}

We examine whether the differences in Table~\ref{tab:matharena_table} are concentrated at a particular difficulty level. Within each benchmark, we rank its 30 questions by Pass@1 over the full trace pool and divide them into six equal bins of five questions. Each row of Table~\ref{tab:difficulty-binned} therefore pools 20 questions, with every benchmark contributing equally, and averages correctness over the same ten seeds as Table~\ref{tab:matharena_table}. Borda-JC uses the same candidate samples and judge evidence. This analysis uses full-pool Pass@1 only to define the bins, not as an input to any aggregation method.

\begin{table}[H]
    \centering
    \small
    \caption{\textbf{MathArena-Full performance stratified by problem difficulty.} Bin 1 contains the five easiest questions from each benchmark and bin 6 the five hardest, as ranked by full-pool Pass@1; each row contains 20 questions in total.  $\Delta_{\mathrm{WSC}}$ and $\Delta_{\mathrm{Borda}}$ denote the accuracy of JC minus WSC and Borda-JC, respectively, in percentage points. Boldface marks a strict row maximum.}
    \small
    \resizebox{\textwidth}{!}{%
        \input{Tables/difficulty_binned.tex}
    }
    \label{tab:difficulty-binned}
\end{table}

The first five bins are largely saturated: JC and WSC differ by at most one percentage point, and neither method has a consistent advantage. In the hardest bin, JC improves over WSC by $2.5$, $4.0$, and $3.0$ points for $\rho=10\%$, $20\%$, and $50\%$, respectively. The comparison with Borda-JC is more mixed at $\rho=10\%$, where Borda-JC is one point higher, but JC leads by $2.0$ and $2.5$ points in the hardest bin at the two larger sampling ratios. The evidence therefore supports a difficulty-dependent benefit rather than uniform improvement across questions.

\subsection{Alternative designs of interaction matrix \texorpdfstring{$\mathbf{J}$}{J}}\label{ssec:alternative-j}
\paragraph{Soft Borda based kernel.} Borda-JC selects the answer maximizing this group-balanced Borda score plus $\mu\sum_{i\in\mathcal{I}_k}\mathbf{h}_i$, with the same $\mu=0.5$ used by JC. It uses the same pairwise and pointwise judgments as JC and requires no additional judge calls.

\paragraph{RBF kernel based interaction.}\label{para:rbf-JC}
We first compute a Gaussian radial basis function (RBF) kernel from the columns of $\mathbf{C}$,
\begin{equation*}
[\mathbf{K}_{\mathrm{RBF}}]_{uv} = \exp\!\left( -\frac{\|\mathbf{C}_{:,u}-\mathbf{C}_{:,v}\|_2^2}{2\ell^2} \right), \qquad \ell>0.
\end{equation*}
For each problem instance, we set the bandwidth using the median heuristic with unit multiplier:
\begin{equation*}
d_{uv}=\|\mathbf{C}_{:,u}-\mathbf{C}_{:,v}\|_2,
\qquad
\ell=\operatorname{median}\{d_{uv}:u<v,\ d_{uv}>0\}.
\end{equation*}
We apply kernel centering with $\mathbf{H}=\mathbf{I}-\frac{1}{N}\mathbf{1}\mathbf{1}^{\top}$, followed by spectral normalization:
\begin{equation*}
\widetilde{\mathbf{K}}_{\mathrm{RBF}}
=\frac{\mathbf{H}\mathbf{K}_{\mathrm{RBF}}\mathbf{H}}
{\lambda_{\max}(\mathbf{H}\mathbf{K}_{\mathrm{RBF}}\mathbf{H})}.
\end{equation*}
The corresponding interaction matrix is
\begin{equation} \label{eq:RBF-J}
\mathbf{J}_{\mathrm{RBF}}
= \mathbf{C}\bigl[(1-\eta)\mathbf{I}+\eta\widetilde{\mathbf{K}}_{\mathrm{RBF}}\bigr]\mathbf{C}^{\top},
\qquad \eta\in[0,1].
\end{equation}
At $\eta=0$, the construction reduces to the Gram matrix $\mathbf{C}\mathbf{C}^{\top}$. For $\eta>0$, the RBF kernel additionally captures interactions between different columns of $\mathbf{C}$, weighted by the similarity of their comparative profiles. Kernel centering removes the common similarity component, while spectral normalization ensures that the kernel has the same operator norm as the identity. The parameter $\eta$ controls the balance between the Gram and RBF terms.

For both MathArena-Full and lower-tier AIME, we use $\mu=0.5$ and $\eta=0.2$. The RBF-JC results are included in Tables~\ref{tab:matharena_table_full} and~\ref{tab:worst_full}. On MathArena-Full, RBF-JC achieves an average accuracy of $97.11\%$, compared with $97.06\%$ for JC. On lower-tier AIME, RBF-JC achieves $62.83\%$, compared with $61.89\%$ for JC and Borda-JC, and obtains the best or tied-best performance in five of the six settings.

\paragraph{MathArena-Full Results (Table~\ref{tab:matharena_table}) with Standard Deviation}\label{ssec:matharena-full-results}
Table~\ref{tab:matharena_table_full} provides the complete version of Table~\ref{tab:matharena_table}, including sample standard deviations over 10 seeds and the RBF-JC variant described in Appendix~\ref{para:rbf-JC}.

\begin{table}[H]
    \centering
    \caption{\textbf{Full MathArena-Full results corresponding to Table~\ref{tab:matharena_table}.} Entries report mean accuracy $\pm$ sample standard deviation over 10 seeds; RBF-JC is included as an additional variant.}
    \footnotesize
    \begin{adjustbox}{max width=\textwidth}
        \begingroup
        \newcommand{\TableOneStd}[1]{\pm#1}
        \newcommand{\TableOneTabcolsep}{2pt}
        \input{Tables/5-1}
        \endgroup
    \end{adjustbox}
    \label{tab:matharena_table_full}
\end{table}

\paragraph{Lower-Tier AIME Results (Table~\ref{tab:worst}) with Standard Deviation}\label{ssec:lower-tier-full-results}
Table~\ref{tab:worst_full} provides the complete version of Table~\ref{tab:worst}, including the standard-deviation statistics under the Table~\ref{tab:worst} reporting convention and the additional RBF-JC variant.

\begin{table}[H]
    \centering
    \caption{\textbf{Full lower-tier AIME results corresponding to Table~\ref{tab:worst}.} Entries report mean accuracy $\pm$  standard-deviation across 10 random seeds, with each seed’s accuracy computed over all 30 problems, used by the Table~\ref{tab:worst} protocol. RBF-JC is included as an additional variant. The final row reports the average over $R$ without aggregating the per-setting standard deviations.}
    \footnotesize
    \begin{adjustbox}{max width=\textwidth}
        \begingroup
        \newcommand{\TableTwoStd}[1]{\pm#1}
        \input{Tables/5-2}

        \endgroup
    \end{adjustbox}
    \label{tab:worst_full}
\end{table}

\paragraph{Ablation of the independent evaluation term.}
We compare Gram-JC and Borda-JC with and without $\mathbf{h}$ on MathArena-Full and in a separate lower-tier AIME experiment. The latter uses the weakest $R\in\{3,5,10,15,20,30\}$ generators with 50\% trace sampling per problem. Within each paired ablation, the sampled traces, pairwise judgments, and interaction construction are held fixed, and the ``without $\mathbf{h}$'' variant only removes $\mu\langle\mathbf{h},\mathbf{x}\rangle$. The full variants use $\mu=0.5$. Table~\ref{tab:h-ablation-j} reports averages over the 12 cells of Table~\ref{tab:matharena_table} and the six values of $R$ in the lower-tier AIME experiment.

\begin{table}[H]
    \centering
    \caption{\textbf{Effect of the independent evaluation term on comparative aggregation.} Entries report mean accuracy (\%); no hyperparameter is retuned after removing $\mathbf{h}$. $\Delta_{\mathbf{h}}$ is the signed change in percentage points from adding $\mathbf{h}$, computed before rounding. Bold denotes the best or tied-best construction in each accuracy column.}
    \label{tab:h-ablation-j}
    \footnotesize
    \input{Tables/jc_h_ablation}
\end{table}

Adding $\mathbf{h}$ improves the average accuracy of both constructions on MathArena-Full, by $12.40$ and $10.69$ percentage points for Gram-JC and Borda-JC, respectively. In the separate lower-tier AIME experiment, adding $\mathbf{h}$ improves Gram-JC by $0.54$ points but reduces Borda-JC by $3.12$ points. This indicates that the fixed $\mu$ need not transfer across comparative scores with different numerical scales.

\subsection{Sensitivity to Pair Order and Prompt Design}\label{ssec:judge-sensitivity}

\paragraph{Pair order.} \label{para:pair-order}
For each pairwise comparison, we query the judge twice with the positions of the two traces swapped. Let $s_{ij}$ be the probability returned when $y_i$ is presented first and $s_{ji}$ the probability returned when $y_j$ is presented first. We use the bidirectionally normalized probabilities
\begin{equation*}
    P_{\theta}(y_i \succeq y_j)
    = \frac{\exp(s_{ij})}{\exp(s_{ij})+\exp(s_{ji})},
    \qquad
    P_{\theta}(y_j \succeq y_i)
    = \frac{\exp(s_{ji})}{\exp(s_{ij})+\exp(s_{ji})}.
\end{equation*}
To evaluate sensitivity to presentation order, we evaluate aggregation in the setting of Table~\ref{tab:matharena_table} using either both directions, only the forward judgment, or only the reverse judgment. For each single-direction condition, the unqueried opposite score is inferred as the complement of the retained probability before applying the same normalization. All conditions use identical sampled traces, independent scores, seeds, and JC objective, and require no additional judge calls.

\begin{table}[H]
    \centering
    \caption{\textbf{Sensitivity of JC to pair presentation order on MathArena-Full.} Entries report mean accuracy $\pm$ sample standard deviation. The bidirectional column reproduces the JC entries in Table~\ref{tab:matharena_table}.}
    \label{tab:pair-order-sensitivity}
    \footnotesize        \input{Tables/pair_order_sensitivity.tex}
\end{table}

Bidirectional querying obtains an accuracy of $97.1\%$, averaged uniformly over the 12 combinations. The corresponding accuracy averaged over the two single-order conditions is $96.8\%$, a difference of $0.22$ percentage points. This indicates that the aggregate result is similar when averaged over the two possible presentation orders in this setting.

\paragraph{Prompt wording and output format.}
We additionally compare three pairwise-judge prompts on lower-tier AIME traces: the original probability prompt, a paraphrased probability prompt, and a direct binary prompt that requests only which response is better. We use a fixed comparison set of 7,840 sampled trace pairs spanning 784 answer pairs, each evaluated in both presentation orders for all three prompts. We consider $R\in\{3,5,10,15,20,30\}$ weakest generators, with 50\%-per-problem sampling, ten repeats, and $\mu=0.5$. The judge model and settings, trace pairs and presentation orders, sampled traces, independent scores, JC objective, and tie handling are held fixed across prompts.
The original probability template is given in Appendix~\ref{ssec:prompt_ising_J}. The complete paraphrased probability and direct binary templates are given in Appendix~\ref{ssec:prompt_ising_J_paraphrased} and~\ref{ssec:prompt_ising_J_binary}, respectively. 

\begin{table}[H]
    \centering
    \caption{\textbf{Sensitivity of JC to pairwise-judge prompt design on lower-tier AIME traces.} Entries report mean accuracy (\%) $\pm$ standard deviation (in percentage points) across 10 random seeds, with each seed’s accuracy computed over all 30 problems. All three prompts use identical trace pairs and aggregation samples.}
    \label{tab:prompt-sensitivity}
    \footnotesize        \input{Tables/prompt_sensitivity.tex}
\end{table}

The accuracies averaged over the six values of $R$ are $64.8\%$, $64.7\%$, and $65.9\%$ for the original probability, paraphrased probability, and direct binary prompts, respectively. The overall accuracies are similar across the three prompts, although individual settings can differ more substantially; for example, at $R=15$, direct binary judging improves over the original probability prompt by $5.3$ percentage points. These results indicate comparable average performance for the three tested prompts under this comparison protocol.

\section{Derivation of the Gram--Borda Decomposition}\label{sec:interpretation}
Fix an answer group $\mathcal{I}_k$ and let $U$ be uniformly distributed over its trace indices. Write $Z_j=\sqrt{p_{Uj}}$, where $p_{ij}=p_\theta(y_i\succeq y_j)$. For $\mathbf{x}=\mathbf{1}_{\mathcal{I}_k}$, the definitions of $\mathbf{C}$ and $\mathbf{J}_{\mathrm{Borda}}$ give
\begin{align*}
\mathbf{x}^{\top}\mathbf{J}\mathbf{x}
&=\|\mathbf{C}^{\top}\mathbf{x}\|_2^2
=\sum_{j=1}^{N}\frac{1}{n_j}\bigl(\mathbb{E}[Z_j]\bigr)^2,\\
\mathbf{x}^{\top}\mathbf{J}_{\mathrm{Borda}}\mathbf{x}
&=\sum_{j=1}^{N}\frac{1}{n_j}\mathbb{E}[Z_j^2].
\end{align*}
Subtracting the two expressions and using $\operatorname{Var}(Z_j)=\mathbb{E}[Z_j^2]-(\mathbb{E}[Z_j])^2$ proves \eqref{eq:gram-borda-relation}. The expectations and variances are over $U$ only, with the preference values held fixed; no answer-level homogeneity assumption is needed.

\section{Prompt Templates}\label{sec:prompts}
\subsection{Prompts Used in Math Reasoning Tasks}
\subsubsection{Ising-h Prompt} \label{ssec:prompt_ising_h}

The prompt to query $\mathbf{h}$ in math reasoning tasks consists of three parts as follows.

\begin{messagebox}{user}
    Please reason step by step, and put your final answer within \verb|\\boxed{}|.
    
\{question\}
\end{messagebox}

\begin{messagebox}{assistant}
\{response (cot+answer)\}
    
\end{messagebox}

\begin{messagebox}{user}
Please evaluate the above answer based on the following criteria: 
\begin{enumerate}
    \item Is the answer correct? 
    \item Is the reasoning process correct? 
\end{enumerate}
Please choose an evaluation score among 0, 0.1, 0.2, 0.3, 0.4, 0.5, 0.6, 0.7, 0.8, 0.9, 1.0.

Please only output the evaluation score.
\end{messagebox}

\subsubsection{Ising-J Prompt}\label{ssec:prompt_ising_J}

\begin{messagebox}{user}
Suppose there are two responses to the same question. Please output the probability that Response 1 is a better answer than Response 2.
\\

\#\#\#\# Question \#\#\#\#

\{question\}\\

\#\#\#\# Response 1 \#\#\#\#

\{response1 (cot+answer)\}\\

\#\#\#\# Response 2 \#\#\#\#

\{response2 (cot+answer)\}\\

\#\#\#\# Instruction \#\#\#\#

Now, please output the probability (a real number between 0 and 1) that Response 1 is a better answer than Response 2. Please only output the number.

\end{messagebox}

\subsubsection{Paraphrased Probability Prompt}\label{ssec:prompt_ising_J_paraphrased}
The following prompt is sent as a single \texttt{user} message with no additional system prompt. Its output is constrained to a single number in $[0,1]$.

\begin{messagebox}{user}
You will be given a question together with two candidate responses. Estimate the probability that Response 1 is the better answer.

\#\#\#\# Question \#\#\#\#

\{question\}

\#\#\#\# Candidate Response 1 \#\#\#\#

\{response1\}

\#\#\#\# Candidate Response 2 \#\#\#\#

\{response2\}

\#\#\#\# Output Instruction \#\#\#\#

Return a single number between 0 and 1 representing the probability that Response 1 is better than Response 2. Output only the number, with no explanation or additional text.
\end{messagebox}

\subsubsection{Direct Binary Prompt}\label{ssec:prompt_ising_J_binary}
The following prompt is sent as a single \texttt{user} message with no additional system prompt. Its output is constrained to exactly \texttt{A} or \texttt{B}.

\begin{messagebox}{user}
Suppose there are two responses to the same question. Choose which response is the better answer.

\#\#\#\# Question \#\#\#\#

\{question\}

\#\#\#\# Response A \#\#\#\#

\{response\_a\}

\#\#\#\# Response B \#\#\#\#

\{response\_b\}

\#\#\#\# Instruction \#\#\#\#

Which response is better? Output exactly one uppercase letter: A or B. Do not output a score, probability, explanation, or any other text.
\end{messagebox}

\subsubsection{Generation Prompt}
\begin{messagebox}{user}
    \{question\}
    
    Please reason step by step, and put your final answer within \verb|\\boxed{}|.
    
\end{messagebox}

\subsection{Prompt Used in Code Reasoning Tasks}\label{ssec:code_reasoning}
\subsubsection{Ising-h Prompt} \label{ssec:prompt_ising_h_code}

The prompt to query $\mathbf{h}$ in code reasoning tasks consists of three parts as follows.
\begin{messagebox}{user}
Given the following Python function and input, predict the output.

Function:
\{code\_str\}

Input:
\{input\_str\}

Please think step by step after "Reasoning:\verb|\n\n|" and then leave the output after "Output:\verb|\n\n|". Note the output should be a python object and please ignore markdown format.\{question\}
\end{messagebox}

\begin{messagebox}{assistant}
    \{response\}
\end{messagebox}

\begin{messagebox}{user}
Please evaluate the above answer based on the following criteria: 
\begin{enumerate}
    \item Is the answer correct? 
    \item Is the reasoning process correct? 
\end{enumerate}
Please choose an evaluation score among 0, 0.1, 0.2, 0.3, 0.4, 0.5, 0.6, 0.7, 0.8, 0.9, 1.0.

Please only output the evaluation score.
\end{messagebox}

\subsubsection{Ising-J Prompt} \label{ssec:prompt_ising_J_code}

\begin{messagebox}{user}
Suppose there are two responses to the same Python function and input. Please output the probability that Response 1 is a better answer than Response 2. \\

\#\#\#\# Python function and input \#\#\#\# \\

Function:\\
\{code\_str\}
\\

Input:\\
\{input\_str\}
\\

\#\#\#\# Response 1 \#\#\#\#\\
\{content1\}
\\

\#\#\#\# Response 2 \#\#\#\#\\
\{content2\}
\\

\#\#\#\# Instruction \#\#\#\#\\

Now, please output the probability (a real number between 0 and 1) that Response 1 is a better answer than Response 2. Please only output the number.
\end{messagebox}

\subsubsection{Generation Prompts}

\begin{messagebox}{user}
Given the following Python function and input, predict the output.

Function:
\{code\_str\}

Input:
\{input\_str\}

Please think step by step after "Reasoning:\verb|\n\n|" and then leave the output after "Output:\verb|\n\n|". Note the output should be a python object and please ignore markdown format.\{question\}
\end{messagebox}


\end{document}

%% file: Tables/5-1_main.tex
\begingroup
\providecommand{\TableOneStd}[1]{}
\providecommand{\TableOneTabcolsep}{4pt}
\setlength{\tabcolsep}{\TableOneTabcolsep}
\begin{tabular}{lrrrrrrrrrrr}
\toprule
Dataset & $\rho$ & Pass@1 & BoN & KT & SC & WSC & Copeland & Elo & \makecell{BT} & Borda-JC & JC \\
\midrule
\multirow{3}{*}{\makecell{AIME\\2025}}
 & 10\% & $70.3\TableOneStd{0.9}$ & $87.6\TableOneStd{2.3}$ & $96.0\TableOneStd{2.1}$ & $94.0\TableOneStd{2.1}$ & $95.3\TableOneStd{1.9}$ & $92.0\TableOneStd{1.9}$ & $83.3\TableOneStd{5.2}$ & $89.7\TableOneStd{4.6}$ & $\bm{96.3}\TableOneStd{2.9}$ & $95.7\TableOneStd{3.2}$ \\
 & 20\% & $69.6\TableOneStd{0.7}$ & $88.4\TableOneStd{1.0}$ & $94.7\TableOneStd{1.7}$ & $93.2\TableOneStd{1.2}$ & $\bm{96.3}\TableOneStd{1.1}$ & $88.3\TableOneStd{0.0}$ & $78.7\TableOneStd{6.5}$ & $90.0\TableOneStd{1.6}$ & $\bm{96.3}\TableOneStd{1.1}$ & $95.7\TableOneStd{1.6}$ \\
 & 50\% & $69.7\TableOneStd{0.4}$ & $88.4\TableOneStd{0.6}$ & $96.0\TableOneStd{1.4}$ & $94.5\TableOneStd{1.6}$ & $\bm{96.7}\TableOneStd{0.0}$ & $88.3\TableOneStd{0.0}$ & $74.3\TableOneStd{4.5}$ & $90.0\TableOneStd{0.0}$ & $\bm{96.7}\TableOneStd{0.0}$ & $\bm{96.7}\TableOneStd{0.0}$ \\
\cmidrule(lr){1-12}
\multirow{3}{*}{\makecell{BRUMO\\2025}}
 & 10\% & $85.8\TableOneStd{1.6}$ & $94.5\TableOneStd{1.4}$ & $98.0\TableOneStd{2.3}$ & $97.3\TableOneStd{2.4}$ & $99.3\TableOneStd{1.4}$ & $95.1\TableOneStd{3.5}$ & $90.3\TableOneStd{4.3}$ & $94.3\TableOneStd{3.5}$ & $\bm{100.0}\TableOneStd{0.0}$ & $\bm{100.0}\TableOneStd{0.0}$ \\
 & 20\% & $86.2\TableOneStd{0.7}$ & $94.3\TableOneStd{1.2}$ & $98.7\TableOneStd{2.3}$ & $98.5\TableOneStd{1.7}$ & $99.7\TableOneStd{1.1}$ & $95.0\TableOneStd{3.3}$ & $90.7\TableOneStd{4.4}$ & $95.0\TableOneStd{2.4}$ & $\bm{100.0}\TableOneStd{0.0}$ & $\bm{100.0}\TableOneStd{0.0}$ \\
 & 50\% & $86.0\TableOneStd{0.4}$ & $94.3\TableOneStd{0.6}$ & $99.7\TableOneStd{1.1}$ & $98.2\TableOneStd{1.7}$ & $\bm{100.0}\TableOneStd{0.0}$ & $90.1\TableOneStd{2.0}$ & $82.0\TableOneStd{3.9}$ & $90.3\TableOneStd{2.9}$ & $\bm{100.0}\TableOneStd{0.0}$ & $\bm{100.0}\TableOneStd{0.0}$ \\
\cmidrule(lr){1-12}
\multirow{3}{*}{\makecell{HMMT\\2025\\(Feb)}}
 & 10\% & $58.4\TableOneStd{1.6}$ & $83.9\TableOneStd{1.9}$ & $90.7\TableOneStd{7.5}$ & $92.7\TableOneStd{2.1}$ & $96.2\TableOneStd{2.9}$ & $80.7\TableOneStd{3.4}$ & $78.7\TableOneStd{5.7}$ & $86.3\TableOneStd{2.5}$ & $96.3\TableOneStd{2.9}$ & $\bm{97.0}\TableOneStd{1.9}$ \\
 & 20\% & $58.1\TableOneStd{1.1}$ & $82.9\TableOneStd{0.8}$ & $94.3\TableOneStd{2.7}$ & $94.5\TableOneStd{1.8}$ & $96.0\TableOneStd{2.6}$ & $80.1\TableOneStd{2.2}$ & $81.7\TableOneStd{3.9}$ & $84.7\TableOneStd{1.7}$ & $96.7\TableOneStd{2.7}$ & $\bm{98.7}\TableOneStd{1.7}$ \\
 & 50\% & $58.3\TableOneStd{0.6}$ & $83.4\TableOneStd{0.8}$ & $96.0\TableOneStd{2.6}$ & $94.0\TableOneStd{1.4}$ & $96.7\TableOneStd{1.6}$ & $78.3\TableOneStd{0.2}$ & $75.3\TableOneStd{6.5}$ & $83.7\TableOneStd{1.1}$ & $97.0\TableOneStd{1.9}$ & $\bm{98.7}\TableOneStd{1.7}$ \\
\cmidrule(lr){1-12}
\multirow{3}{*}{\makecell{HMMT\\2025\\(Nov)}}
 & 10\% & $87.4\TableOneStd{1.8}$ & $88.5\TableOneStd{3.1}$ & $90.0\TableOneStd{4.2}$ & $93.5\TableOneStd{1.5}$ & $93.8\TableOneStd{2.4}$ & $84.1\TableOneStd{6.4}$ & $84.7\TableOneStd{7.2}$ & $83.7\TableOneStd{6.2}$ & $\bm{94.3}\TableOneStd{2.7}$ & $93.0\TableOneStd{2.9}$ \\
 & 20\% & $87.8\TableOneStd{1.1}$ & $89.4\TableOneStd{0.9}$ & $92.7\TableOneStd{3.1}$ & $93.5\TableOneStd{0.5}$ & $94.0\TableOneStd{1.4}$ & $80.0\TableOneStd{6.3}$ & $79.7\TableOneStd{4.6}$ & $78.3\TableOneStd{6.3}$ & $94.3\TableOneStd{1.6}$ & $\bm{95.0}\TableOneStd{2.4}$ \\
 & 50\% & $87.7\TableOneStd{0.5}$ & $89.6\TableOneStd{0.7}$ & $94.0\TableOneStd{2.1}$ & $93.3\TableOneStd{0.0}$ & $\bm{94.3}\TableOneStd{1.6}$ & $68.3\TableOneStd{3.5}$ & $69.3\TableOneStd{5.8}$ & $68.0\TableOneStd{3.2}$ & $\bm{94.3}\TableOneStd{1.6}$ & $\bm{94.3}\TableOneStd{1.6}$ \\
\bottomrule
\end{tabular}
\endgroup

%% file: Tables/5-2_main.tex
\begingroup
\providecommand{\TableTwoStd}[1]{}
\setlength{\tabcolsep}{2.2pt}
\begin{tabular}{ccccccccccc}
\toprule
$R$ & Pass@1 & BoN & KT & SC & WSC & Copeland & Elo & BT & Borda-JC & JC \\
\midrule
3 & $11.2\TableTwoStd{2.2}$ & $23.7\TableTwoStd{4.3}$ & $9.3\TableTwoStd{1.4}$ & $19.6\TableTwoStd{3.8}$ & $23.5\TableTwoStd{4.2}$ & $26.7\TableTwoStd{4.4}$ & $24.3\TableTwoStd{4.3}$ & $\bm{27.0}\TableTwoStd{4.4}$ & $\bm{27.0}\TableTwoStd{4.4}$ & $\bm{27.0}\TableTwoStd{4.4}$ \\
5 & $16.0\TableTwoStd{2.6}$ & $31.8\TableTwoStd{4.7}$ & $13.6\TableTwoStd{0.9}$ & $29.5\TableTwoStd{4.5}$ & $31.8\TableTwoStd{4.6}$ & $34.3\TableTwoStd{4.8}$ & $32.0\TableTwoStd{4.7}$ & $\bm{36.3}\TableTwoStd{4.8}$ & $\bm{36.3}\TableTwoStd{4.8}$ & $\bm{36.3}\TableTwoStd{4.8}$ \\
10 & $26.9\TableTwoStd{3.2}$ & $50.0\TableTwoStd{5.0}$ & $40.9\TableTwoStd{3.7}$ & $43.0\TableTwoStd{4.9}$ & $51.6\TableTwoStd{4.9}$ & $59.3\TableTwoStd{4.9}$ & $46.3\TableTwoStd{5.0}$ & $\bm{60.0}\TableTwoStd{4.9}$ & $58.0\TableTwoStd{4.9}$ & $59.7\TableTwoStd{4.9}$ \\
15 & $34.3\TableTwoStd{3.2}$ & $63.1\TableTwoStd{4.8}$ & $57.9\TableTwoStd{4.4}$ & $54.7\TableTwoStd{4.9}$ & $68.7\TableTwoStd{4.6}$ & $69.2\TableTwoStd{4.4}$ & $48.7\TableTwoStd{5.0}$ & $\bm{76.3}\TableTwoStd{4.3}$ & $74.0\TableTwoStd{4.4}$ & $76.0\TableTwoStd{4.3}$ \\
20 & $40.9\TableTwoStd{3.2}$ & $67.3\TableTwoStd{4.7}$ & $69.4\TableTwoStd{3.7}$ & $71.5\TableTwoStd{4.4}$ & $77.7\TableTwoStd{4.2}$ & $70.8\TableTwoStd{4.4}$ & $52.7\TableTwoStd{5.0}$ & $70.7\TableTwoStd{4.6}$ & $\bm{84.3}\TableTwoStd{3.6}$ & $82.3\TableTwoStd{3.8}$ \\
30 & $52.1\TableTwoStd{3.0}$ & $76.5\TableTwoStd{4.2}$ & $83.6\TableTwoStd{2.6}$ & $83.3\TableTwoStd{3.7}$ & $88.2\TableTwoStd{3.2}$ & $63.7\TableTwoStd{4.5}$ & $44.7\TableTwoStd{5.0}$ & $60.7\TableTwoStd{4.9}$ & $\bm{91.7}\TableTwoStd{2.8}$ & $90.0\TableTwoStd{3.0}$ \\
\midrule
\textbf{Average} & $30.23$ & $52.07$ & $45.78$ & $50.27$ & $56.91$ & $54.00$ & $41.44$ & $55.17$ & $\bm{61.89}$ & $\bm{61.89}$ \\
\bottomrule
\end{tabular}
\endgroup

%% file: Tables/cost_table.tex
\begin{tabular}{ccccccccc}
\toprule
N & \multicolumn{2}{c}{0.20} & \multicolumn{2}{c}{0.40} & \multicolumn{2}{c}{0.60} & \multicolumn{2}{c}{0.80} \\
\cmidrule(lr){2-3} \cmidrule(lr){4-5} \cmidrule(lr){6-7} \cmidrule(lr){8-9}
$\kappa$ & JC ($\mathbf{h}\equiv\bm{0}$) & WSC ($\mathbf{J}\equiv\bm{0}$)  &  JC ($\mathbf{h}\equiv\bm{0}$) & WSC ($\mathbf{J}\equiv\bm{0}$)  &  JC ($\mathbf{h}\equiv\bm{0}$) & WSC ($\mathbf{J}\equiv\bm{0}$)  &  JC ($\mathbf{h}\equiv\bm{0}$) & WSC ($\mathbf{J}\equiv\bm{0}$)  \\
\midrule
2 & $44.0 \% / \$0.09$ & $54.4 \% / \$0.20$ & $51.9 \% / \$0.17$ & $62.8 \% / \$0.41$ & $56.9 \% / \$0.25$ & ${\color{blue}65.0} \% / \$0.63$ & $60.6 \% / \$0.34$ & ${\color{blue}65.0} \% / \$0.83$ \\
3 & $42.9 \% / \$0.25$ & $57.2 \% / \$0.22$ & $52.5 \% / \$0.48$ & $63.7 \% / \$0.41$ & $58.8 \% / \$0.74$ & $63.7 \% / \$0.63$ & $60.0 \% / \$1.00$ & $62.5 \% / \$0.83$ \\
4 & \cellcolor{green!20} $66.2 \% / \$0.47$ & $63.1 \% / \$0.22$ &  \cellcolor{green!20}  $75.6 \% / \$0.94$ & $62.8 \% / \$0.42$ &  \cellcolor{green!20} $78.8 \% / \$1.39$ & $64.4 \% / \$0.63$ &  \cellcolor{green!20}  $85.6 \% / \$1.85$ & $62.5 \% / \$0.83$ \\
5 &  \cellcolor{green!20}  $68.1 \% / \$0.72$ & $59.1 \% / \$0.21$ &  \cellcolor{green!20}  $78.1 \% / \$1.49$ & $62.5 \% / \$0.42$ &  \cellcolor{green!20}  $80.0 \% / \$2.21$ & $64.4 \% / \$0.64$ &  \cellcolor{green!20}  $85.6 \% / \$2.95$ & $63.1 \% / \$0.83$ \\
6 & $60.6 \% / \$1.11$ & $57.5 \% / \$0.21$ &  \cellcolor{green!20} $75.0 \% / \$2.23$ & $64.4 \% / \$0.42$ &  \cellcolor{green!20} $80.0 \% / \$3.35$ & $63.1 \% / \$0.62$ &  \cellcolor{green!20} $79.4 \% / \$4.47$ & ${\color{blue}65.0} \% / \$0.83$ \\
8 & $62.5 \% / \$2.15$ & $57.5 \% / \$0.22$ & $  \cellcolor{green!20}  {\color{red}80.0} \% / \$4.23$ & $62.8 \% / \$0.41$ & \cellcolor{green!20} $81.2 \% / \$6.37$ & ${\color{blue}65.0} \% / \$0.64$ &  \cellcolor{green!20}  ${\color{red}86.2} \% / \$8.43$ & $63.7 \% / \$0.83$ \\
10 &  \cellcolor{green!20}  ${\color{red}70.0} \% / \$3.36$ & $58.3 \% / \$0.21$ &  \cellcolor{green!20} $75.6 \% / \$6.71$ & $63.4 \% / \$0.42$ &  \cellcolor{green!20}  ${\color{red}83.1} \% / \$10.12$ & $64.4 \% / \$0.61$ &  \cellcolor{green!20}  $85.6 \% / \$13.44$ & $63.1 \% / \$0.84$ \\
12 & $61.9 \% / \$4.66$ & $59.4 \% / \$0.20$ &  \cellcolor{green!20} $76.9 \% / \$9.36$ & $63.1 \% / \$0.41$ &  \cellcolor{green!20}  $80.0 \% / \$14.19$ & $63.7 \% / \$0.61$ &  \cellcolor{green!20}  $80.6 \% / \$18.87$ & $63.1 \% / \$0.83$ \\
14 & $58.8 \% / \$5.76$ & $59.4 \% / \$0.21$ &  \cellcolor{green!20}  $74.4 \% / \$11.52$ & $64.7 \% / \$0.42$ &  \cellcolor{green!20}  $71.2 \% / \$17.58$ & $63.7 \% / \$0.63$ &  \cellcolor{green!20}  $67.5 \% / \$23.35$ & $63.7 \% / \$0.83$ \\
16 & $61.3 \% / \$6.85$ & $60.9 \% / \$0.20$ &  \cellcolor{green!20}  $71.2 \% / \$13.73$ & $63.7 \% / \$0.41$ &  \cellcolor{green!20}  $65.0 \% / \$20.96$ & $64.4 \% / \$0.62$ & $64.4 \% / \$27.79$ & $62.8 \% / \$0.84$ \\
18 & $63.1 \% / \$8.23$ & $61.6 \% / \$0.21$ &  \cellcolor{green!20}  $68.1 \% / \$16.43$ & $63.7 \% / \$0.43$ &  \cellcolor{green!20}  $70.6 \% / \$25.35$ & $63.7 \% / \$0.63$ & \cellcolor{green!20}  $65.6 \% / \$33.53$ & $64.4 \% / \$0.83$ \\
20 & $53.8 \% / \$9.50$ & $59.1 \% / \$0.21$ &  \cellcolor{green!20}  $70.0 \% / \$18.93$ & $60.6 \% / \$0.42$ &  \cellcolor{green!20}  $66.9 \% / \$29.29$ & $63.7 \% / \$0.63$ & $63.1 \% / \$38.75$ & $62.5 \% / \$0.83$ \\
\bottomrule
\end{tabular}

%% file: Tables/difficulty_binned.tex
\begingroup
\setlength{\tabcolsep}{5pt}
\begin{tabular}{ccrrrrrr}
\toprule
$\rho$ & Difficulty bin & Full-pool Pass@1 & WSC & Borda-JC & JC & $\Delta_{\mathrm{WSC}}$ & $\Delta_{\mathrm{Borda}}$ \\
\midrule
\multirow{6}{*}{10\%}
 & 1 (easiest) & 97.1 & 100.0 & 100.0 & 100.0 & $+0.0$ & $+0.0$ \\
 & 2 & 90.5 & 100.0 & 100.0 & 100.0 & $+0.0$ & $+0.0$ \\
 & 3 & 83.7 & 100.0 & 100.0 & 100.0 & $+0.0$ & $+0.0$ \\
 & 4 & 77.6 & 99.5 & 99.5 & 99.5 & $+0.0$ & $+0.0$ \\
 & 5 & 68.0 & 99.5 & 99.5 & 98.5 & $-1.0$ & $-1.0$ \\
 & 6 (hardest) & 35.8 & 78.0 & \textbf{81.5} & 80.5 & $+2.5$ & $-1.0$ \\
\midrule
\multirow{6}{*}{20\%}
 & 1 (easiest) & 97.1 & 100.0 & 100.0 & 100.0 & $+0.0$ & $+0.0$ \\
 & 2 & 90.5 & 100.0 & 100.0 & 100.0 & $+0.0$ & $+0.0$ \\
 & 3 & 83.7 & 100.0 & 100.0 & 100.0 & $+0.0$ & $+0.0$ \\
 & 4 & 77.6 & 99.0 & 99.0 & \textbf{100.0} & $+1.0$ & $+1.0$ \\
 & 5 & 68.0 & 100.0 & 100.0 & 100.0 & $+0.0$ & $+0.0$ \\
 & 6 (hardest) & 35.8 & 80.0 & 82.0 & \textbf{84.0} & $+4.0$ & $+2.0$ \\
\midrule
\multirow{6}{*}{50\%}
 & 1 (easiest) & 97.1 & 100.0 & 100.0 & 100.0 & $+0.0$ & $+0.0$ \\
 & 2 & 90.5 & 100.0 & 100.0 & 100.0 & $+0.0$ & $+0.0$ \\
 & 3 & 83.7 & 100.0 & 100.0 & 100.0 & $+0.0$ & $+0.0$ \\
 & 4 & 77.6 & 100.0 & 100.0 & 100.0 & $+0.0$ & $+0.0$ \\
 & 5 & 68.0 & 100.0 & 100.0 & 100.0 & $+0.0$ & $+0.0$ \\
 & 6 (hardest) & 35.8 & 81.5 & 82.0 & \textbf{84.5} & $+3.0$ & $+2.5$ \\
\bottomrule
\end{tabular}
\endgroup

%% file: Tables/5-1.tex
\begingroup
\providecommand{\TableOneStd}[1]{}
\providecommand{\TableOneTabcolsep}{4pt}
\setlength{\tabcolsep}{\TableOneTabcolsep}
\begin{tabular}{lrrrrrrrrrrrr}
\toprule
Dataset & $\rho$ & Pass@1 & BoN & KT & SC & WSC & Copeland & Elo & \makecell{BT} & Borda-JC & RBF-JC & JC \\
\midrule
\multirow{3}{*}{\makecell{AIME\\2025}}
 & 10\% & $70.3\TableOneStd{0.9}$ & $87.6\TableOneStd{2.3}$ & $96.0\TableOneStd{2.1}$ & $94.0\TableOneStd{2.1}$ & $95.3\TableOneStd{1.9}$ & $92.0\TableOneStd{1.9}$ & $83.3\TableOneStd{5.2}$ & $89.7\TableOneStd{4.6}$ & $\bm{96.3}\TableOneStd{2.9}$ & $95.7\TableOneStd{3.2}$ & $95.7\TableOneStd{3.2}$ \\
 & 20\% & $69.6\TableOneStd{0.7}$ & $88.4\TableOneStd{1.0}$ & $94.7\TableOneStd{1.7}$ & $93.2\TableOneStd{1.2}$ & $\bm{96.3}\TableOneStd{1.1}$ & $88.3\TableOneStd{0.0}$ & $78.7\TableOneStd{6.5}$ & $90.0\TableOneStd{1.6}$ & $\bm{96.3}\TableOneStd{1.1}$ & $96.0\TableOneStd{2.1}$ & $95.7\TableOneStd{1.6}$ \\
 & 50\% & $69.7\TableOneStd{0.4}$ & $88.4\TableOneStd{0.6}$ & $96.0\TableOneStd{1.4}$ & $94.5\TableOneStd{1.6}$ & $\bm{96.7}\TableOneStd{0.0}$ & $88.3\TableOneStd{0.0}$ & $74.3\TableOneStd{4.5}$ & $90.0\TableOneStd{0.0}$ & $\bm{96.7}\TableOneStd{0.0}$ & $\bm{96.7}\TableOneStd{0.0}$ & $\bm{96.7}\TableOneStd{0.0}$ \\
\cmidrule(lr){1-13}
\multirow{3}{*}{\makecell{BRUMO\\2025}}
 & 10\% & $85.8\TableOneStd{1.6}$ & $94.5\TableOneStd{1.4}$ & $98.0\TableOneStd{2.3}$ & $97.3\TableOneStd{2.4}$ & $99.3\TableOneStd{1.4}$ & $95.1\TableOneStd{3.5}$ & $90.3\TableOneStd{4.3}$ & $94.3\TableOneStd{3.5}$ & $\bm{100.0}\TableOneStd{0.0}$ & $\bm{100.0}\TableOneStd{0.0}$ & $\bm{100.0}\TableOneStd{0.0}$ \\
 & 20\% & $86.2\TableOneStd{0.7}$ & $94.3\TableOneStd{1.2}$ & $98.7\TableOneStd{2.3}$ & $98.5\TableOneStd{1.7}$ & $99.7\TableOneStd{1.1}$ & $95.0\TableOneStd{3.3}$ & $90.7\TableOneStd{4.4}$ & $95.0\TableOneStd{2.4}$ & $\bm{100.0}\TableOneStd{0.0}$ & $\bm{100.0}\TableOneStd{0.0}$ & $\bm{100.0}\TableOneStd{0.0}$ \\
 & 50\% & $86.0\TableOneStd{0.4}$ & $94.3\TableOneStd{0.6}$ & $99.7\TableOneStd{1.1}$ & $98.2\TableOneStd{1.7}$ & $\bm{100.0}\TableOneStd{0.0}$ & $90.1\TableOneStd{2.0}$ & $82.0\TableOneStd{3.9}$ & $90.3\TableOneStd{2.9}$ & $\bm{100.0}\TableOneStd{0.0}$ & $\bm{100.0}\TableOneStd{0.0}$ & $\bm{100.0}\TableOneStd{0.0}$ \\
\cmidrule(lr){1-13}
\multirow{3}{*}{\makecell{HMMT\\2025\\(Feb)}}
 & 10\% & $58.4\TableOneStd{1.6}$ & $83.9\TableOneStd{1.9}$ & $90.7\TableOneStd{7.5}$ & $92.7\TableOneStd{2.1}$ & $96.2\TableOneStd{2.9}$ & $80.7\TableOneStd{3.4}$ & $78.7\TableOneStd{5.7}$ & $86.3\TableOneStd{2.5}$ & $96.3\TableOneStd{2.9}$ & $\bm{97.0}\TableOneStd{1.9}$ & $\bm{97.0}\TableOneStd{1.9}$ \\
 & 20\% & $58.1\TableOneStd{1.1}$ & $82.9\TableOneStd{0.8}$ & $94.3\TableOneStd{2.7}$ & $94.5\TableOneStd{1.8}$ & $96.0\TableOneStd{2.6}$ & $80.1\TableOneStd{2.2}$ & $81.7\TableOneStd{3.9}$ & $84.7\TableOneStd{1.7}$ & $96.7\TableOneStd{2.7}$ & $\bm{98.7}\TableOneStd{1.7}$ & $\bm{98.7}\TableOneStd{1.7}$ \\
 & 50\% & $58.3\TableOneStd{0.6}$ & $83.4\TableOneStd{0.8}$ & $96.0\TableOneStd{2.6}$ & $94.0\TableOneStd{1.4}$ & $96.7\TableOneStd{1.6}$ & $78.3\TableOneStd{0.2}$ & $75.3\TableOneStd{6.5}$ & $83.7\TableOneStd{1.1}$ & $97.0\TableOneStd{1.9}$ & $98.3\TableOneStd{1.8}$ & $\bm{98.7}\TableOneStd{1.7}$ \\
\cmidrule(lr){1-13}
\multirow{3}{*}{\makecell{HMMT\\2025\\(Nov)}}
 & 10\% & $87.4\TableOneStd{1.8}$ & $88.5\TableOneStd{3.1}$ & $90.0\TableOneStd{4.2}$ & $93.5\TableOneStd{1.5}$ & $93.8\TableOneStd{2.4}$ & $84.1\TableOneStd{6.4}$ & $84.7\TableOneStd{7.2}$ & $83.7\TableOneStd{6.2}$ & $\bm{94.3}\TableOneStd{2.7}$ & $93.7\TableOneStd{2.5}$ & $93.0\TableOneStd{2.9}$ \\
 & 20\% & $87.8\TableOneStd{1.1}$ & $89.4\TableOneStd{0.9}$ & $92.7\TableOneStd{3.1}$ & $93.5\TableOneStd{0.5}$ & $94.0\TableOneStd{1.4}$ & $80.0\TableOneStd{6.3}$ & $79.7\TableOneStd{4.6}$ & $78.3\TableOneStd{6.3}$ & $94.3\TableOneStd{1.6}$ & $\bm{95.0}\TableOneStd{2.4}$ & $\bm{95.0}\TableOneStd{2.4}$ \\
 & 50\% & $87.7\TableOneStd{0.5}$ & $89.6\TableOneStd{0.7}$ & $94.0\TableOneStd{2.1}$ & $93.3\TableOneStd{0.0}$ & $\bm{94.3}\TableOneStd{1.6}$ & $68.3\TableOneStd{3.5}$ & $69.3\TableOneStd{5.8}$ & $68.0\TableOneStd{3.2}$ & $\bm{94.3}\TableOneStd{1.6}$ & $\bm{94.3}\TableOneStd{1.6}$ & $\bm{94.3}\TableOneStd{1.6}$ \\
\bottomrule
\end{tabular}
\endgroup

%% file: Tables/5-2.tex
\begingroup
\providecommand{\TableTwoStd}[1]{}
\setlength{\tabcolsep}{2.2pt}
\begin{tabular}{cccccccccccc}
\toprule
$R$ & Pass@1 & BoN & KT & SC & WSC & Copeland & Elo & BT & Borda-JC & RBF-JC & JC \\
\midrule
3 & $11.2\TableTwoStd{2.2}$ & $23.7\TableTwoStd{4.3}$ & $9.3\TableTwoStd{1.4}$ & $19.6\TableTwoStd{3.8}$ & $23.5\TableTwoStd{4.2}$ & $26.7\TableTwoStd{4.4}$ & $24.3\TableTwoStd{4.3}$ & $\bm{27.0}\TableTwoStd{4.4}$ & $\bm{27.0}\TableTwoStd{4.4}$ & $\bm{27.0}\TableTwoStd{4.4}$ & $\bm{27.0}\TableTwoStd{4.4}$ \\
5 & $16.0\TableTwoStd{2.6}$ & $31.8\TableTwoStd{4.7}$ & $13.6\TableTwoStd{0.9}$ & $29.5\TableTwoStd{4.5}$ & $31.8\TableTwoStd{4.6}$ & $34.3\TableTwoStd{4.8}$ & $32.0\TableTwoStd{4.7}$ & $\bm{36.3}\TableTwoStd{4.8}$ & $\bm{36.3}\TableTwoStd{4.8}$ & $\bm{36.3}\TableTwoStd{4.8}$ & $\bm{36.3}\TableTwoStd{4.8}$ \\
10 & $26.9\TableTwoStd{3.2}$ & $50.0\TableTwoStd{5.0}$ & $40.9\TableTwoStd{3.7}$ & $43.0\TableTwoStd{4.9}$ & $51.6\TableTwoStd{4.9}$ & $59.3\TableTwoStd{4.9}$ & $46.3\TableTwoStd{5.0}$ & $\bm{60.0}\TableTwoStd{4.9}$ & $58.0\TableTwoStd{4.9}$ & $\bm{60.0}\TableTwoStd{4.9}$ & $59.7\TableTwoStd{4.9}$ \\
15 & $34.3\TableTwoStd{3.2}$ & $63.1\TableTwoStd{4.8}$ & $57.9\TableTwoStd{4.4}$ & $54.7\TableTwoStd{4.9}$ & $68.7\TableTwoStd{4.6}$ & $69.2\TableTwoStd{4.4}$ & $48.7\TableTwoStd{5.0}$ & $76.3\TableTwoStd{4.3}$ & $74.0\TableTwoStd{4.4}$ & $\bm{77.7}\TableTwoStd{4.2}$ & $76.0\TableTwoStd{4.3}$ \\
20 & $40.9\TableTwoStd{3.2}$ & $67.3\TableTwoStd{4.7}$ & $69.4\TableTwoStd{3.7}$ & $71.5\TableTwoStd{4.4}$ & $77.7\TableTwoStd{4.2}$ & $70.8\TableTwoStd{4.4}$ & $52.7\TableTwoStd{5.0}$ & $70.7\TableTwoStd{4.6}$ & $84.3\TableTwoStd{3.6}$ & $\bm{84.7}\TableTwoStd{3.6}$ & $82.3\TableTwoStd{3.8}$ \\
30 & $52.1\TableTwoStd{3.0}$ & $76.5\TableTwoStd{4.2}$ & $83.6\TableTwoStd{2.6}$ & $83.3\TableTwoStd{3.7}$ & $88.2\TableTwoStd{3.2}$ & $63.7\TableTwoStd{4.5}$ & $44.7\TableTwoStd{5.0}$ & $60.7\TableTwoStd{4.9}$ & $\bm{91.7}\TableTwoStd{2.8}$ & $91.3\TableTwoStd{2.8}$ & $90.0\TableTwoStd{3.0}$ \\
\midrule
\textbf{Average} & $30.23$ & $52.07$ & $45.78$ & $50.27$ & $56.91$ & $54.00$ & $41.44$ & $55.17$ & $61.89$ & $\bm{62.83}$ & $61.89$ \\
\bottomrule
\end{tabular}
\endgroup

%% file: Tables/jc_h_ablation.tex
\begin{tabular}{lrrrrrr}
\toprule
& \multicolumn{3}{c}{Table~\ref{tab:matharena_table}} & \multicolumn{3}{c}{Lower-tier AIME} \\
\cmidrule(lr){2-4}\cmidrule(lr){5-7}
Comparative component
& w/o $\mathbf{h}$ & w/ $\mathbf{h}$ & $\Delta_{\mathbf{h}}$
& w/o $\mathbf{h}$ & w/ $\mathbf{h}$ & $\Delta_{\mathbf{h}}$ \\
\midrule
Gram-JC  & $84.66$ & $\bm{97.06}$ & $+12.40$ & $64.29$ & $\bm{64.83}$ & $+0.54$ \\
Borda-JC & $\bm{86.17}$ & $96.86$ & $+10.69$ & $\bm{65.96}$ & $62.83$ & $-3.12$ \\
\bottomrule
\end{tabular}

%% file: Tables/pair_order_sensitivity.tex
\begingroup
\setlength{\tabcolsep}{3.5pt}
\begin{tabular}{llccc}
\toprule
Dataset & $\rho$ & Bidirectional & Forward only & Reverse only \\
\midrule
\multirow{3}{*}{\makecell{AIME\\2025}}
 & 10\% & $95.7\pm3.2$ & $93.3\pm1.6$ & $97.3\pm3.4$ \\
 & 20\% & $95.7\pm1.6$ & $93.7\pm1.1$ & $98.7\pm1.7$ \\
 & 50\% & $96.7\pm0.0$ & $96.3\pm1.1$ & $96.7\pm0.0$ \\
\midrule
\multirow{3}{*}{\makecell{BRUMO\\2025}}
 & 10\% & $100.0\pm0.0$ & $99.7\pm1.1$ & $100.0\pm0.0$ \\
 & 20\% & $100.0\pm0.0$ & $100.0\pm0.0$ & $100.0\pm0.0$ \\
 & 50\% & $100.0\pm0.0$ & $100.0\pm0.0$ & $100.0\pm0.0$ \\
\midrule
\multirow{3}{*}{\makecell{HMMT\\2025\\(Feb)}}
 & 10\% & $97.0\pm1.9$ & $95.3\pm2.3$ & $96.0\pm3.1$ \\
 & 20\% & $98.7\pm1.7$ & $98.7\pm1.7$ & $98.7\pm1.7$ \\
 & 50\% & $98.7\pm1.7$ & $98.0\pm2.3$ & $98.3\pm1.8$ \\
\midrule
\multirow{3}{*}{\makecell{HMMT\\2025\\(Nov)}}
 & 10\% & $93.0\pm2.9$ & $91.0\pm3.2$ & $93.7\pm3.7$ \\
 & 20\% & $95.0\pm2.4$ & $93.0\pm1.9$ & $96.7\pm2.2$ \\
 & 50\% & $94.3\pm1.6$ & $93.7\pm1.1$ & $95.3\pm1.7$ \\
\bottomrule
\end{tabular}
\endgroup

%% file: Tables/prompt_sensitivity.tex
\begingroup
\setlength{\tabcolsep}{4pt}
\begin{tabular}{rccc}
\toprule
$R$ & Original probability & Paraphrased probability & Direct binary \\
\midrule
3  & $26.3\pm4.4$ & $26.3\pm4.4$ & $26.3\pm4.4$ \\
5  & $36.3\pm4.8$ & $36.3\pm4.8$ & $36.3\pm4.8$ \\
10 & $61.3\pm4.9$ & $63.3\pm4.8$ & $63.7\pm4.8$ \\
15 & $80.0\pm4.0$ & $78.0\pm4.1$ & $85.3\pm3.5$ \\
20 & $89.7\pm3.0$ & $88.3\pm3.2$ & $89.0\pm3.1$ \\
30 & $95.3\pm2.1$ & $96.0\pm2.0$ & $94.5\pm2.3$ \\
\bottomrule
\end{tabular}
\endgroup